\documentclass[10pt,twocolumn,letterpaper]{article}

\usepackage{iccv}
\usepackage{times}
\usepackage{epsfig}
\usepackage{graphicx}
\usepackage{amsmath}
\usepackage{amssymb}
\usepackage{booktabs}
\usepackage{caption}
\usepackage{colortbl}
\usepackage{cuted}
\usepackage{longtable}
\usepackage{multirow, makecell}
\usepackage{tabularx}
\usepackage[table,xcdraw]{xcolor}
\usepackage{xspace}
\usepackage{bbding}
\usepackage{xcolor}
\usepackage{lipsum}  
\usepackage{relsize}
\usepackage[export]{adjustbox}
\usepackage[toc,page]{appendix}
\usepackage{calc}

\definecolor{maroon}{HTML}{6E0A1E}
\definecolor{dorange}{HTML}{3b2000} 

\usepackage[pagebackref=true,breaklinks=true,letterpaper=true,colorlinks,bookmarks=false,citecolor=orange]{hyperref}
\hypersetup{linkcolor=maroon}
\hypersetup{urlcolor=orange}

\usepackage{cleveref}

\makeatletter
\renewcommand{\paragraph}{%
  \@startsection{paragraph}{4}%
  {\z@}{.3ex \@plus 1ex \@minus .2ex}{-1em}%
  {\normalfont\normalsize\bfseries\color{dorange}}%
}
\makeatother

\makeatletter
\newcommand{\oneptsmaller}[1]{%
  \begingroup
  \fontsize{\dimexpr\f@size pt-1pt}{\f@baselineskip}\selectfont
  #1%
  \endgroup
}
\makeatother

\usepackage{pifont}
\usepackage{adjustbox}
\newcolumntype{R}[2]{%
    >{\adjustbox{angle=#1}\bgroup}%
    l%
    <{\egroup}%
}
\newcommand{\nvideos}{18,619\xspace}
\newcommand{\ncategories}{50\xspace}
\newcommand{\setfont}{\ttfamily}
\newcommand{\testset}{{\setfont test}\xspace}
\newcommand{\trainset}{{\setfont train}\xspace}
\newcommand{\traintrainset}{{\setfont train-known}\xspace}
\newcommand{\traintestset}{{\setfont train-unseen}\xspace}
\newcommand{\testtrainset}{{\setfont test-known}\xspace}
\newcommand{\testtestset}{{\setfont test-unseen}\xspace}

\newcommand{\bh}{\mathbf{h}}
\newcommand{\bx}{\mathbf{x}}

\newcommand{\br}{\mathbf{r}}
\newcommand{\bz}{\mathbf{z}}
\newcommand{\bu}{\mathbf{u}}

\newcommand{\method}[1]{{\oneptsmaller{\sffamily #1}}}

\newcommand{\cmark}{\checkmark}%
\newcommand{\xmark}{\scalebox{0.85}{\ding{53}}}%
\newcommand{\TheDataset}{Common Objects in 3D\xspace}
\newcommand{\TheDatasetAbbrev}{CO3D\xspace}
\newcommand{\TheMethod}{NerFormer\xspace}
\newcommand{\bb}[1]{{\textbf{\underline{#1}}}}
\renewcommand{\b}[1]{{\underline{#1}}}
\newcommand{\transformerAnerfAopacityAdimdown}{\textbf{\TheMethod}}
\newcommand{\nerfAopacity}{\method{NeRF\cite{mildenhall2020nerf}} }
\newcommand{\nerfAautodecAopacity}{\method{NeRF+AD} }
\newcommand{\nerfAviewpoolAopacity}{\method{NeRF+WCE}\cite{henzler2021unsupervised}}
\newcommand{\nvAautodec}{\method{NV\cite{lombardi2019neural}}}
\newcommand{\nvAviewpool}{\method{NV+WCE}}
\newcommand{\idr}{\method{IDR\cite{yariv2020multiview}}}
\newcommand{\idrAautodec}{\method{IDR+AD}}
\newcommand{\srnAharmonicAautodecAopacity}{\method{SRN+$\gamma$}}
\newcommand{\srnAautodecAopacity}{\method{SRN\cite{sitzmann19scene}}}
\newcommand{\srnAharmonicAviewpoolAopacity}{\method{SRN+WCE+$\gamma$}}
\newcommand{\srnAviewpoolAopacity}{\method{SRN+WCE}}
\newcommand{\dvrAautodec}{\method{DVR\cite{niemeyer2020differentiable}}}
\newcommand{\dvrAharmonicAautodec}{\method{DVR+$\gamma$}}
\newcommand{\softrasAautodec}{\method{P3DMesh\cite{ravi2020pytorch3d}}}
\newcommand{\softrasAviewpool}{\method{P3DMesh}}
\newcommand{\pointcloudsAautodecAopacities}{\method{IPC}}
\newcommand{\pointcloudsAviewpoolAopacities}{\method{IPC+WCE}}
\DeclareMathOperator*{\argmin}{arg\,min}

\newlength\myheightinline
\newlength\mydepthinline
\settototalheight\myheightinline{Xygp}
\newcommand*\inlinegraphics[1]{%
  \settototalheight\myheightinline{Xygp}%
  \settodepth\mydepthinline{Xygp}%
  \raisebox{-\mydepthinline}{\includegraphics[height=\myheightinline]{#1}}%
}

\Crefname{equation}{Eq.}{Eqs.}
\Crefname{figure}{Fig.}{Figs.}
\Crefname{table}{Tab.}{Tabs.}
\Crefname{tabular}{Tab.}{Tabs.}
\Crefname{section}{Sec.}{Secs.}
\Crefname{appendix}{Sec.}{Secs.}

\crefname{equation}{eq.}{eqs.}
\crefname{figure}{fig.}{figs.}
\crefname{tabular}{tab.}{tabs.}
\crefname{section}{sec.}{secs.}
\crefname{table}{tab.}{tabs.}
\crefname{appendix}{sec.}{secs.}

\iccvfinalcopy 
\def\arxivsubmission

\def\iccvPaperID{6639} 

\ifx\arxivsubmission\undefined
\ificcvfinal\fi
\else
\fi

\title{%
\TheDataset%
:%
\\
{\Large Large-Scale Learning and Evaluation of Real-life 3D Category Reconstruction}\\
}%
\author{%
Jeremy Reizenstein$^{1}$
\and
Roman Shapovalov$^{1}$ \vspace{.025cm}
\and
Philipp Henzler$^{2}$ \vspace{.025cm}
\and
Luca Sbordone$^{1}$ \vspace{.025cm}
\and
Patrick Labatut$^{1}$
\and
David Novotny$^{1}$
\vspace{.1cm}
\and
{\tt\small \{reizenstein,romansh,lsbordone,plabatut,dnovotny\}@fb.com}
\and
{\tt\small \{p.henzler\}@cs.ucl.ac.uk}
\vspace{.1cm}
\and 
\hspace{1.5cm} 
$^{1}$Facebook AI Research \and
$^{2}$University College London \and
\vspace{.1cm}
\and
{\Large\inlinegraphics{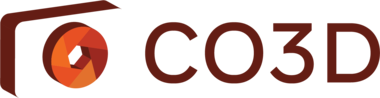}}\hspace{0.5cm}%
\texttt{\textbf{\normalsize\href{https://github.com/facebookresearch/co3d}{https://github.com/facebookresearch/co3d}}}%
}

\begin{document}

\twocolumn[{%
\renewcommand\twocolumn[1][]{#1}%
\maketitle
\vspace{-0.6cm}%
\begin{center}
    \centering
    \includegraphics[width=\textwidth]{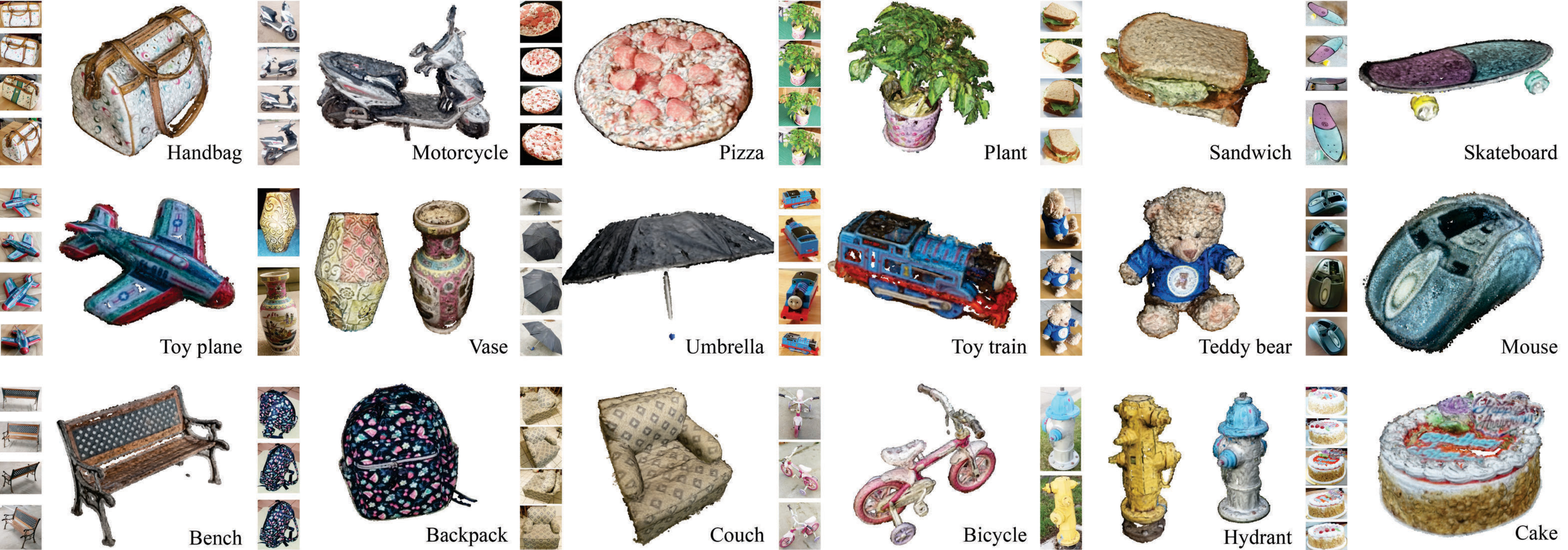}
    \captionof{figure}{We introduce the \textbf{\TheDataset (\TheDatasetAbbrev)} dataset comprising 1.5 million multi-view images of almost 19k objects from 50 MS-COCO categories annotated with accurate cameras and 3D point clouds (visualized above). 
    }\label{f:splash}
\end{center}%
}]

\begin{abstract}\vspace{-1.0em}
Traditional approaches for learning 3D object categories have been predominantly trained and evaluated on synthetic datasets due to the unavailability of real 3D-annotated category-centric data.
Our main goal is to facilitate advances in this field by collecting real-world data in a magnitude similar to the existing synthetic counterparts.
The principal contribution of this work is thus a large-scale dataset, called \emph{\TheDataset}, with real multi-view images of object categories annotated with camera poses and ground truth 3D point clouds.
The dataset contains a total of 1.5 million frames from nearly 19,000 videos capturing objects from 50 MS-COCO categories and, as such, it is significantly larger than alternatives both in terms of the number of categories and objects.

We exploit this new dataset to conduct one of the first large-scale ``in-the-wild'' evaluations of several new-view-synthesis and category-centric 3D reconstruction methods. 
Finally, we contribute \emph{\TheMethod}~- a novel neural rendering method that leverages the powerful Transformer to reconstruct an object given a small number of its views.
\end{abstract}

\section{Introduction}\label{s:intro}

Recently, the community witnessed numerous advances in deeply learning to reconstruct category-centric 3D models.
While a large variety of technical approaches was proposed \cite{sitzmann19scene,tulsiani2017learning,dosovitskiy16generating,henzler19escaping,henzler2021unsupervised,niemeyer19occupancy,chen2019learning}, they are predominantly trained and benchmarked either on synthetic data  \cite{chang2015shapenet}, or on real datasets of specific object categories such as birds \cite{tulsiani2017learning} or chairs \cite{lpt2013ikea}.
The latter is primarily a consequence of a lack of relevant real-world datasets with 3D ground truth.  

Our main goal is therefore to collect a large-scale open real-life dataset of common objects in the wild annotated with 3D ground truth.
While the latter can be collected with specialized hardware (turn-table 3D scanner, dome \cite{joo17panoptic}), it is challenging to reach the scale of synthetic datasets \cite{chang2015shapenet} comprising thousands of instances of diverse categories.

Instead, we devise a photogrammetric approach only requiring object-centric multi-view RGB images.
Such data can be effectively gathered in huge quantities by means of crowd-sourcing ``turn-table'' videos captured with smartphones, which are nowadays a commonly owned accessory.
The mature Structure-from-Motion (SfM) framework then provides 3D annotations by tracking cameras and reconstructs a dense 3D point cloud capturing the object surface.

To this end, we collected almost 19,000 videos of 50 MS-COCO categories with 1.5 million frames, each annotated with camera pose, where 20\% of the videos are annotated with a semi-manually verified high-resolution 3D point cloud.
As such, the dataset exceeds alternatives \cite{choi2016large,ahmadyan2020objectron,ignition2020google} in terms of number of categories and objects.

Our work is an extension of the dataset from \cite{henzler2021unsupervised}.
Here, we significantly increase the dataset size from less than 10 categories to 50 and, more importantly, conduct a human-in-the-loop check ensuring reliable accuracy of all cameras.
Finally, the dataset from \cite{henzler2021unsupervised} did not contain any point cloud annotations, the examples of which are in \cref{f:splash}.

We also propose a novel \emph{\TheMethod} model that, given a small number of input source views, learns to reconstruct object categories in our dataset.
\TheMethod mates
two of the main workhorses of machine learning and 3D computer vision: Transformers \cite{vaswani2017attention} and neural implicit rendering \cite{mildenhall2020nerf}.
Specifically, given a set of 3D points along a rendering ray, features are sampled from known images and stacked into a tensor.
The latter is in fact a ray-depth-ordered sequence of sets of sampled features which admits processing with a sequence-to-sequence Transformer.
Therefore, by means of alternating feature pooling attention and ray-wise attention layers, \TheMethod learns to jointly aggregate features from the source views and raymarch over them.

Importantly, \TheMethod\ outperforms a total of 14 baselines which leverage the most common shape representations to date. 
As such, our paper conducts one of the first truly large-scale evaluations of learning 3D object categories in the wild.

\section{Related Work}\label{s:related}

In this section we review current 3D datasets and related methods in the areas of single-image reconstruction, generative modelling and novel-view synthesis.
\begin{table}[th!]
    \centering
    \small
    \setlength{\tabcolsep}{3.4pt}
    \begin{tabular}{lrrrrrrr}
        Dataset&
        \cite{chang2015shapenet}&
        \cite{xiang14beyond}&
        \cite{choi2016large}&
        \cite{ahmadyan2020objectron}&
        \cite{ignition2020google}&
        \cite{henzler2021unsupervised}&
        \textbf{Ours}\\
        \toprule
        Real &
        \xmark&
        \cmark&
        \cmark&
        \cmark&
        \cmark&
        \cmark&
        \cmark\\
        \# Categories &
        55 &
        12&
        9&
        9 &
        NA &
        7 &
        50 \\
        \# Objects &
        51k&
        36k&
        2k&
        15k&
        2k&
        2k&
        19k\\
        3D GT&
        full &
        approx. &
        depth &
        pcl, box &
        full &
        pcl &
        pcl \\
        Multi-view &
        full &
        none &
        full&
        limited &
        full&
        full&
        full \\\bottomrule
    \end{tabular}
    \caption{\TheDataset\ compared to alternatives. 
    The ``pcl" abbreviation stands for ``point clouds".}
    \label{t:dataset_comparison}
\end{table}

\paragraph{3D object datasets}
The main enabler of early works in 3D reconstruction was the synthetic ShapeNet \cite{chang2015shapenet} dataset. 
Pascal3D~\cite{xiang14beyond} introduced a real world dataset providing pose estimation for images, but only approximate 3D models. 
Choi et al.~\cite{choi2016large} provide a large set of realobject-centric RGB-D videos, however only a small subset is annotated with 3D models and cameras.
Increasing the number of categories and objects, Objectron~\cite{ahmadyan2020objectron} contains object-centric videos, object/camera poses, point clouds, surface planes and 3D bounding boxes. Unfortunately, only a limited number of object-centric videos cover full 360 degrees. 
Our dataset further increases the number of categories by a factor of 5 and covers the full 360 degree range. 
Requiring 3D scanners, GSO~\cite{ignition2020google} provides clean full 3D models including textures of real world objects. Due to the requirement of 3D scanning, it contains less objects.
A detailed comparison of the aforementioned datasets is presented in tab.~\ref{t:dataset_comparison}.

\paragraph{3D reconstruction} A vast amount of methods studied fully supervised 3D reconstruction from 2D images making use of several different representations: voxel grids \cite{choy20163d,girdhar2016learning}, meshes \cite{gkioxari19mesh,wang2018pixel2mesh}, point clouds \cite{fan2017point, yang2019pointflow}, signed distance fields, \cite{park2019deepsdf,atzmon2020sal} or continuous occupancy fields \cite{mescheder2019occupancy,chen2019learning, genova2019learning,genova2019deep}. 

Methods overcoming the need for 3D supervision are based on differentiable rendering allowing for comparison of 2D images rather than 3D shapes \cite{rezende2016unsupervised, tulsiani2017multi, kar2017learning}. 
Generating images from meshes was achieved via soft rasterization in \cite{kato2018neural,liu19soft,tulsiani2018multi,kanazawa18learning,Chen2019,li2020self,zhang2020image,goel2020shape,wang2018pixel2mesh}. 
Volumetric represenations are projected to 2D via differentiable raymarching \cite{henzler19escaping,gadelha20173d,lombardi2019neural,mildenhall2020nerf,liu2020neural,schwarz2020graf} or in a similar fashion via sphere tracing for signed distance fields \cite{niemeyer2020differentiable,yariv2020multiview}. 
\cite{insafutdinov2018unsupervised} introduce differentiable point clouds. 
Another line of work focuses on \emph{neural rendering}, i.e. neural networks are trained to approximate the rendering function \cite{nguyen-phuoc19hologan,sitzmann19scene}.
\cite{kulkarni19canonical,kulkarni20articulation-aware} map pixels to object-specific derformable template shapes.
\cite{novotny17learning,novotny18capturing} canonically align point clouds of object-categories in an unsupervised fashion, but do not reason about colour. Exploiting symmetries and reasoning about lighting \cite{wu2020unsupervised}, compose appearance, shape and lighting.

Similar to us, \cite{saito19pifu,yu2020pixelnerf,henzler2021unsupervised,trevithick2020grf} utilize per-pixel warp-conditioned embedding \cite{henzler2021unsupervised}.
In contrast to our method, multi-view aggregation is handled by averaging over encodings which is prone to noise. Instead, we learn the aggregation by introducing the \TheMethod module. 
Finally, a very recent IBRNet \cite{wang2021ibrnet} learns to copy existing colors from known views in an IBR fashion \cite{hedman18deepblending,buehler2001unstructured}, whereas our method can hallucinate new colors, which is crucial since we leverage far fewer source views (at most 9).

\begin{figure*}[th!]
\centering
\includegraphics[width=\linewidth]{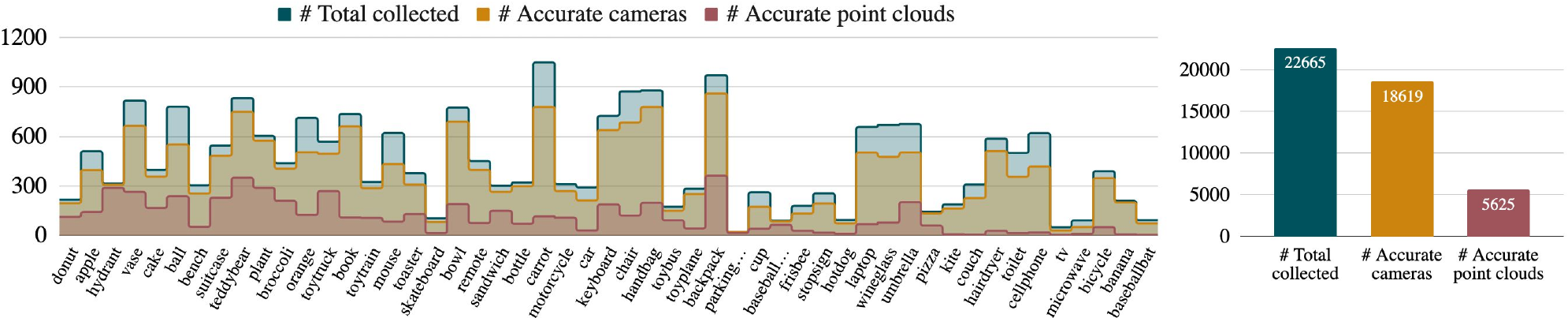}%
\vspace{-0.2cm}%
\caption{%
\textbf{Statistics of the \TheDataset dataset} reporting the total (right) and per-category (left) numbers of collected videos, and the number of videos with accurate cameras and point clouds.%
\label{t:amt_pointclouds_stats}%
}\vspace{-0.2cm}%
\end{figure*} 
\section{\TheDataset}\label{s:data_3dgt}

In this section we describe the dataset collection process.

\paragraph{AMT video collection}
In order to scale the collection of object-centric videos, we crowd-sourced it on Amazon Mechanical Turk (AMT).
Each AMT task asks a worker to select an object of a given category, place it on a solid surface and take a video where they keep the whole object in view while moving full circle around it.
We pre-selected 50 MS-COCO~\cite{lin14microsoft} categories (listed in \cref{t:amt_pointclouds_stats}) comprising stationary objects that are typically large enough to be reconstructed.
The workers were instructed to avoid actions that would hinder the ensuing reconstruction stage 
such as abrupt movements leading to motion blur.
Each video was reviewed to ensure that it fulfills the requirements.

\begin{figure*}[th]\centering%
\includegraphics[width=0.95\linewidth]{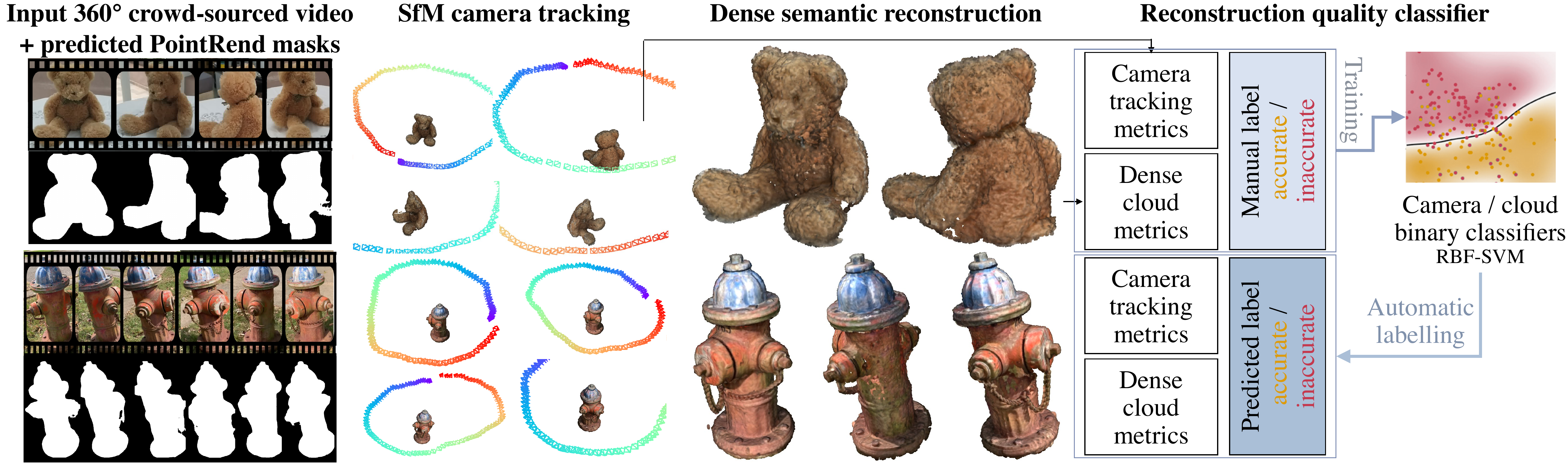}%
\vspace{-0.2cm}%
\caption{\textbf{3D ground truth for \TheDataset} was generated with active learning.
Videos are first annotated with cameras and dense point clouds with COLMAP \cite{schoenberger2016sfm}.
Given several reconstruction metrics and manual binary annotations (``accurate''/``inaccurate'') of a representative subset of reconstructions, we train an SVM that automatically labels all videos.\label{f:amt_pointclouds}}%
\vspace{-0.2cm}%
\end{figure*}

\paragraph{Generating 3D ground-truth}
As explained below, we use off-the-shelf software to produce object masks, camera tracking, and 3D reconstructions for each video. We then semi-automatically filter out poor reconstructions.

\paragraph{1) Sparse reconstruction}
Given the set of valid object-centric videos, we reconstruct the extrinsic (3D location and orientation) and intrinsic (calibration) properties of the cameras that captured the videos.
To this end, each video is first converted into a time-ordered sequence of images $\mathcal{V} = ( I^i \mid I^i \in \mathbb{R}^{3 \times H \times W} )_{i=1}^{N_I} $ by extracting $n_I=100$ frames uniformly spaced in time.
The frames are then fed to the COLMAP SfM pipeline \cite{schoenberger2016sfm} which annotates each image with 
camera projection matrices $\mathcal{P} = (P^i \mid P^i \in \mathbb{R}^{4 \times 4} )_{i=1}^{N_I}$.
\Cref{f:amt_pointclouds} shows example camera tracks together with estimated sparse scene geometries.

\paragraph{2) Object segmentation}%
We segment the object in each image $I^i$ with PointRend  \cite{kirillov2020pointrend}, state-of-the-art instance segmentation method, resulting in a sequence of soft binary masks $\mathcal{M} = (M^i \mid M^i \in [0, 1]^{H \times W})_{i=1}^{N_I}$ per video.
Note that, while masks aid further dense reconstruction, we have not used them for camera tracking which is typically anchored to the background regions.

\paragraph{3) Semantic dense reconstruction}
Having obtained the camera motions and segmentations, we now describe the process of annotating the captured objects with a 3D surface.
We first execute the multi-view stereo (MVS) algorithm of COLMAP \cite{schoenberger2016mvs} to generate per-frame dense depth maps $(D^i \mid D^i \in \mathbb{R}_+^{H \times W} )_{i=1}^{N_I}$.
We then run COLMAP's point cloud fusion algorithm, which back-projects the depth values masked with $\mathcal{M}$ and retains the points that are consistent across frames, to get a point cloud $\mathcal{P}(\mathcal{V}) = \{\bx_i\}_{i=1}^{N_\mathcal{P}}$.
Example dense point clouds are visualized in \cref{f:amt_pointclouds}.

\paragraph{4) Labelling reconstruction quality with Human-in-the-loop.} 
Since our reconstruction pipeline is completely automated, any of the aforementioned steps can fail, resulting in unreliable 3D annotations.
We thus incorporate a semi-manual check that filters inaccurate reconstructions.

To this end, we employ an active learning \cite{cohn1994improving} pipeline which cycles between: a) manually labelling the point-cloud and camera-tracking quality; b) retraining a ``quality"  SVM classifier; and c) automatically estimating the shape/tracking quality of unlabelled videos. 
Details of this process are deferred to the supplementary.

\paragraph{5) The dataset}
The active-learned SVM aids the final dataset filtering step.
As accurate camera annotations are crucial for the majority of recent 3D category reconstruction methods, the first filtering stage completely removes all scenes with camera tracking classified as ``inaccurate" (18\% of all videos).
While all videos that pass the camera check are suitable for training, the scenes that pass both camera and point cloud checks (30\% of the videos with accurate cameras) comprise the pool from which evaluation videos are selected.
Note, a failure to pass the point cloud check does not entail that the corresponding scene is irreconstructible and should therefore be removed from training --
instead this merely implies that the MVS method~\cite{schoenberger2016mvs} failed, while other alternatives (see \cref{s:exp}) could succeed.

\Cref{t:amt_pointclouds_stats} summarizes the size of \TheDataset.
Reconstructing a single video took on average 1h 56 minutes with the majority of execution time spent on GPU-accelerated MVS and correspondence estimation.
This amounts to the total of 43819 GPU-hours of reconstruction time distributed to a large GPU cluster.
\section{Learning 3D categories in the wild}\label{s:method}

Here, we describe the problem set, give an overview of implicit shape representations, and explain our main technical contribution: the \TheMethod\ neural rendering model.

\paragraph{Problem setting}
We tackle the task of generating a representation of appearance and geometry of an object given a small number of its observed source RGB views 
$\{ I_i^\text{src} \}_{i=1}^{N_\text{src}}$ and their cameras $\{ P_i^\text{src} \}_{i=1}^{N_\text{src}}$ which, in our case, are samples from the set of frames $\mathcal{V}_v$ and cameras $\mathcal{P}_v$ of a video $v$. 

In order to visualize the generated shape and appearance, we differentiably render it to a \emph{target} view for which only the camera parameters 
$P^\text{tgt} \in \mathbb{R}^{4 \times 4}$ are known.
This results in a render $\hat I^\text{tgt}$ which is, during training, incentivised to match the ground truth target view  $I^\text{tgt}$.

Methods are trained on a dataset of category-centric videos $\{\mathcal{V}_v\}_{v=1}^{n_\mathcal{V}}$, which allows to exploit the regular structure of the category for better generalization to previously unseen objects.
Note that, in our experiments (\cref{s:exp}), we additionally consider ``overfitting" with $n_\mathcal{V}=1$.

\subsection{Representing common objects in 3D} \label{s:method_shape_representations} %
In order to predict novel views, all methods need to reconstruct 3D shape in some form.
The latter can be represented with a plethora of recent frameworks such as 
voxel grids \cite{tulsiani17multi-view,lombardi2019neural}, 
implicit surfaces \cite{mildenhall2020nerf,niemeyer19occupancy,niemeyer2020differentiable}, 
point clouds \cite{insafutdinov2018unsupervised, wiles2020synsin}, 
or meshes \cite{liu19soft,ravi2020pytorch3d}.
Among those, implicit representations have been successfully applied to reconstructing real scenes and object categories~\cite{mildenhall2020nerf,henzler19escaping,yu2020pixelnerf}, therefore we choose to use them in our pipeline. 

\paragraph{Implicit surface}
An implicit surface is defined as a level set $\mathcal{S}_f = \{ \bx \mid f(\bx, \bz) = C, C \in \mathbb{R} \}$ of a function 
$f: \mathbb{R}^3 \times \mathbb{D}_z \mapsto \mathbb{R}$
that accepts a 3D point 
$\bx \in \mathbb{R}^3$ and $D_z$-dimensional latent code $\bz \in \mathbb{R}^{D_z}$.
In addition to $f$, which represents geometry, a second function 
$c: \mathbb{R}^3 \times \mathbb{S}^2 \times \mathbb{D}_z \mapsto \mathbb{R}^3$ assigns colors $c(\bx, \br, \bz)$ to the input points $\bx$.
Note that, in line with recent work \cite{mildenhall2020nerf,yariv2020multiview}, $c$ is further conditioned on the 3D direction vector $\br \in \mathbb{S}^2$ from which $\bx$ is imaged in order to model viewpoint-dependent effects such as specularities.
Finally, both functions $f$ and $c$ depend on the latent code $\bz$ encoding geometry and appearance of the scene.
Changes in $\bz$ alter the level set of $f$ and colors $c$ allowing for representing different instances of object categories.

Following recent successes of~\cite{schwarz2020graf,henzler2021unsupervised}, 
we model category-specific 3D shapes with opacity functions $f_o$. 
Specifically, $f_o$ assigns $f_o(\bx, \bz) = 0$ to the unoccupied points  $\bx \notin \mathcal{S}_f$, and $f(\bx', \bz)>0$ to surface points $\bx' \in \mathcal{S}_f$.

\paragraph{Neural implicit surfaces} %
Recent methods implement functions $f_o$ and $c$ as multi-layer perceptrons (MLP) $f_\text{MLP}$ and $c_\text{MLP}$ \cite{mildenhall2020nerf,niemeyer2020differentiable,ha16hypernetworks,atzmon2020sal}.
They typically learn shallow specialized networks on top of the shared deep feature extractor
$f'_\text{MLP}$: $f_\text{MLP} = f_\text{HEAD} \circ f'_\text{MLP}$ and 
$c_\text{MLP} = c_\text{HEAD} \circ f'_\text{MLP}$.

We depart from representing occupancies with plain MLPs as they process input 3D points $\bx$ independently, without any form of spatial reasoning, which is crucial in our case where input source views provide only a partial information about the reconstructed shape.

\paragraph{Positional embedding}\label{s:method_boosting_positional}
Following \cite{mildenhall2020nerf,vaswani17attention}, avoiding loss of detail, we pre-process the raw 3D coordinates $\bx$ with a positional embedding (PE)
$\gamma(\bx)
= [\sin(\bx), \cos(\bx), ..., \sin(2^{N_f} \bx), \cos(2^{N_f} \bx)] \in \mathbb{R}^{2 N_f} 
$
before feeding to $f_\text{MLP}(\bx, \bz)$.
While \cite{mildenhall2020nerf} was the first to demonstrate benefits of PE, in \cref{s:exp} we also combine PE with other pre-NeRF methods, such as SRN \cite{sitzmann19scene} or DVR \cite{niemeyer2020differentiable}.

\paragraph{Rendering an implicit surface}\label{s:rendering}
In order to admit image-supervised learning, implicit surfaces are converted into an explicit representation of appearance and geometry with a rendering function $r$.
Formally, given a target camera $P^\text{tgt}$, 
the goal is to generate the target image $I^\text{tgt} = r(f, c, P^\text{tgt})$ which depicts the scene from $P^\text{tgt}$'s viewpoint.

We render opacity fields with the Emission-Absorption model (EA).
EA renders the RGB value 
$I^{\text{tgt}}_\bu \in \mathbb{R}^3$ at a pixel 
$\bu \in \{1, \dots, W\} \times \{1, ..., H\}$, 
by evaluating the opacity function $f_o(\bx, \bz)$ for an ordered point-tuple $(\bx_i^{\br_\bu})_{i=1}^{N_\text{S}}$ sampled along $\bu$'s projection ray $\br_\bu$ 
at approximately equidistant intervals $\Delta$.
The color 
$
I^\text{tgt}_\bu(\br_\bu, \bz) = 
\sum_{i=1}^{N_\text{S}} 
w_i(\bx_i^{\br_\bu}, \bz) 
c(\bx_i^{\br_\bu}, \br_\bu, \bz)
$
is a sum of per-point colors $c(\bx_i^{\br_\bu}, \br_\bu, \bz)$ weighted by the emission-absorption product $w_i = \left(\prod_{j=0}^{i-1} T_j\right) \left(1 - T_i\right)$
with $T_i = \exp(-\Delta f_o(\bx_i^{\br_u}, \bz))$.

\subsection{Latent shape encoding $\bz$}
A crucial part of a category-centric reconstructor is the latent embedding $\bz$.
Early methods \cite{girdhar2016learning,wu16learning,tatarchenko2016multi,tulsiani2018multi,liu19soft} predicted a global scene encoding $\bz_\text{global} = \Phi_\text{CNN}(I^\text{src})$ with a deep convolutional network $\Phi_\text{CNN}$ that solely analyzed the colors of source image pixels.
While this approach was successful for the synthetic ShapeNet dataset where shapes are rigidly aligned, it has been recently shown in \cite{henzler2021unsupervised,yu2020pixelnerf} that such approach is infeasible in real world settings where objects are arbitrarily placed in the 3D space.
This is because, unlike in the former case where color-based shape inference is possible since similar images of aligned scenes generate similar 3D shapes, in the latter case, similarly looking images of unaligned scenes can generate vastly different 3D.

\begin{figure}[t]
\centering
\includegraphics[width=\linewidth]{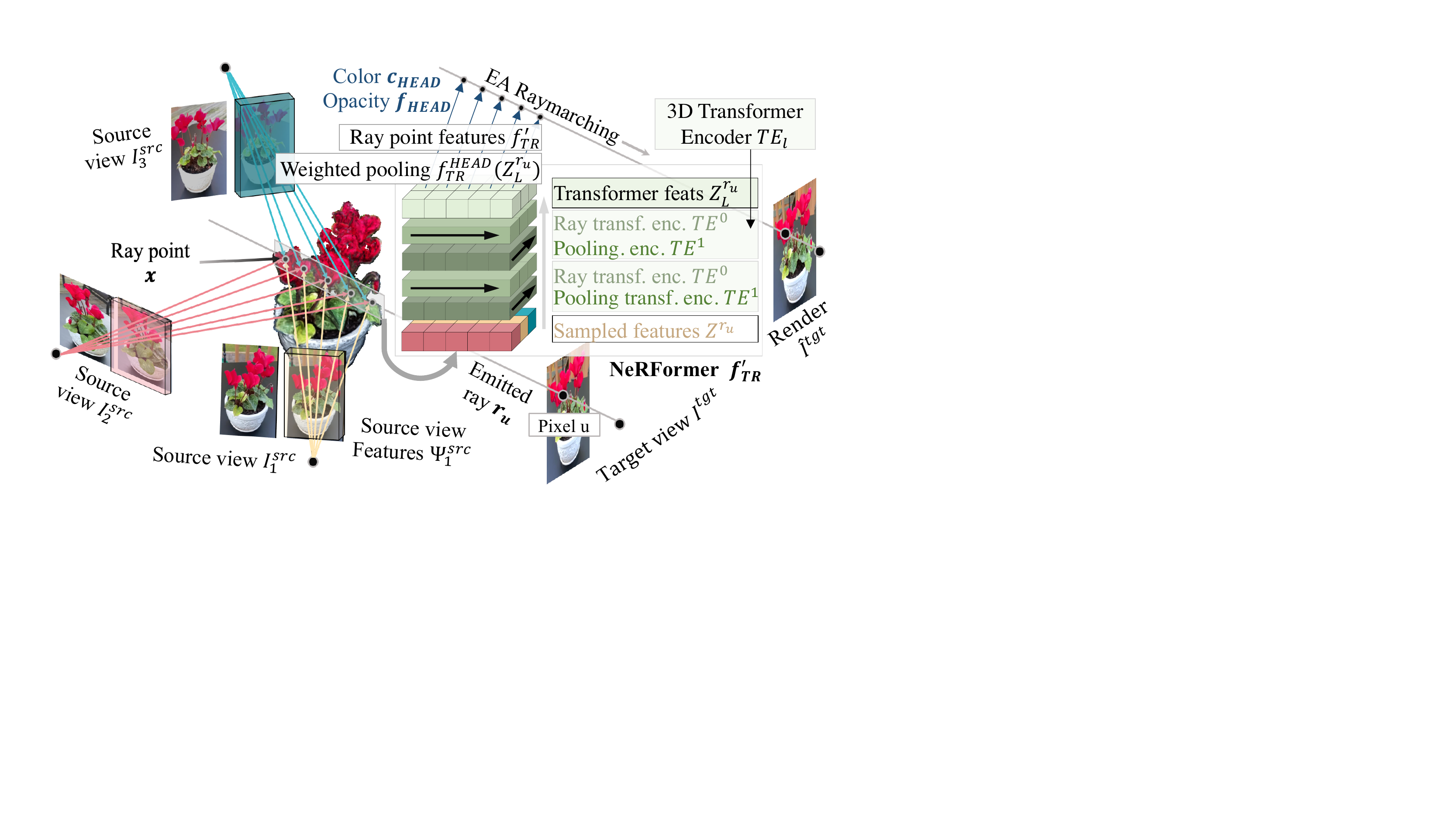}%
\vspace{-0.3cm}%
\caption{%
\textbf{We propose \TheMethod} which jointly learns to pool features from source views and to raymarch by means of a series of transformers alternating between attention along the ray and pooling dimensions.%
\label{f:nerformer}}%
\vspace{-0.2cm}%
\end{figure}%

\begin{figure}%
\input{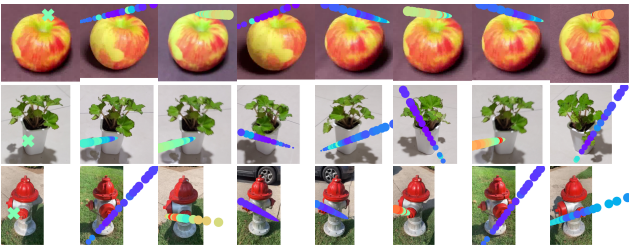}%
\vspace{-0.5cm}%
\caption{%
\textbf{\TheMethod\ learns to attend to features from source images}.
A ray is emitted from a target-image pixel (1st column), and its points are projected to the source views (columns 2-8) from which features are sampled.
For each source feature, \TheMethod\ predicts attention (red=high, blue=low) for aggregating to a single source embedding per point.
Note how the attention model implicitly learns to pool features from nearby source views.%
\label{f:nerformer_attention}}%
\end{figure}

\paragraph{Warp-conditioned embedding} \label{s:method_boosting_wce}%
To fix the latter, 
\cite{henzler2021unsupervised}
proposed Warp-Conditioned Embedding (WCE):
given a world coordinate point $\bx$ and a source view $I^\text{src}$ with camera $P^\text{src}$, 
the warp-conditioned embedding $\bz_\text{WCE} \in \mathbb{R}^{D_z}$
$$
\bz_\text{WCE}(\bx, I^\text{src}, P^\text{src})
=
\Psi_\text{CNN}(I^\text{src}) [\pi_{P^\text{src}}(\bx)],
$$
is formed by sampling a tensor of source image descriptors 
$\Psi_\text{CNN}(I^\text{src}) \in \mathbb{R}^{D_z \times H \times W}$ 
at a 2D location $\pi_{P^\text{src}}(\bx)$.
Here, $\pi_{P^\text{src}}(\bx) = P^\text{src} [\bx; 1] = d_\bu [\bu; 1]$ expresses perspective camera projection of a 3D point $\bx$ to a pixel $\bu$ with depth $d_\bu \in \mathbb{R}$.
Intuitively, since $\bz_\text{WCE}$ is a function of the world coordinate $\bx$, the ensuing implicit $f$ can perceive the specific 3D location of the sampled appearance element in the world coordinates, which in turn enables $f$ to learn invariance to rigid scene misalignment. 

In the common case where multiple source views are given,
the aggregate WCE $\bz_\text{WCE}^*(\bx, \{I_i^\text{src}\}, \{P_i^\text{src}\})$ is defined as a concatenation of the mean and standard deviation of the set of view-specific source embeddings 
$\{ \bz_\text{WCE}(\bx,I_i^\text{src}, P^\text{src}) \}_{i=1}^{N_\text{src}}$.

\paragraph{Boosting baselines with WCE}
The vast majority of existing methods for learning 3D categories leverage global shape embeddings $\bz_\text{global}$, which renders them inapplicable to our real dataset.
As WCE has been designed to alleviate this critical flaw, and because of its generic nature, in this paper we endow state-of-the-art category-centric 3D reconstruction methods with WCE in order to enable them for learning category-specific models on our real dataset.

To this end, we complement SRN \cite{sitzmann19scene}, NeuralVolumes \cite{lombardi2019neural}, and the Implicit Point Cloud (discussed later) with WCE.
These extensions are detailed in the supplementary.

\subsection{Attention is all you nerf}\label{s:method_nerformer}%

\paragraph{Limitations of WCE} While \cite{henzler2021unsupervised} has demonstrated that the combination of NeRF and WCE (termed NeRF-WCE) leads to performance improvements, our experiments indicated that a major shortcoming of NeRF-WCE is its inability to deal with cases where parts of the 3D domain are labelled with noisy WCE.
This is because NeRF's MLP $f'_\text{MLP}$ independently processes each 3D point and, as such, cannot detect failures and recover from them via \emph{spatial reasoning}.
The 3D deconvolutions of Neural Volumes \cite{lombardi2019neural} are a potential solution, but we found that the method ultimately produces blurry renders due to the limited resolution of the voxel grid.
The LSTM marcher of SRN \cite{sitzmann19scene} is capable of spatial reasoning, which is however somewhat limited due to the low-capacity of the LSTM cell. 
Last but not least, a fundamental flaw is that simple averaging of the source-view WCE embeddings can suppress important features.

\begin{figure*}[ht]
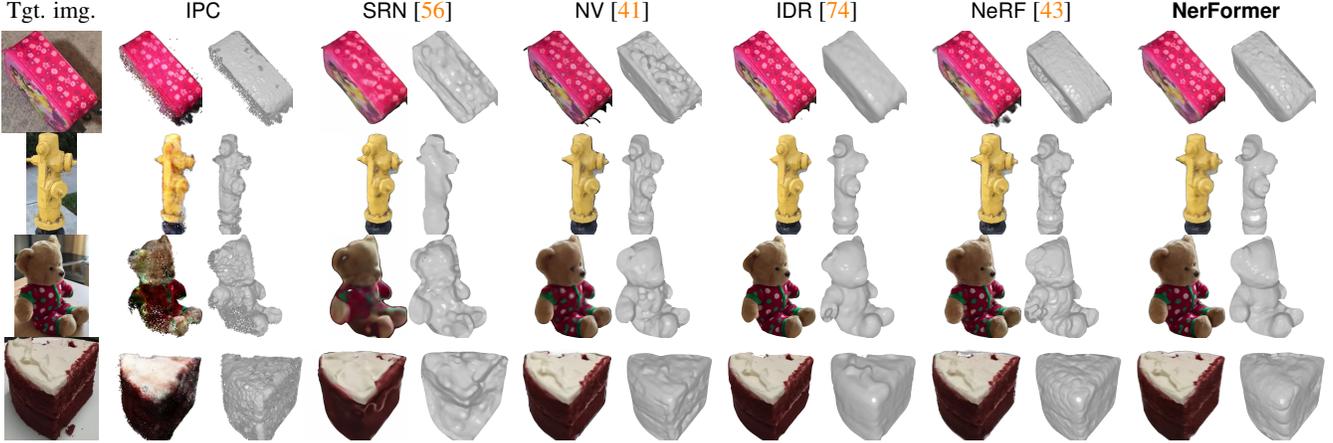

\centering%
\graphicspath{{figures/qual_overfit/}}
\newcommand{\methcolimw}{1.36cm}%
\newcommand{\methcolimh}{3.0cm}%
\newcommand{\methodcol}[3]{%
\includegraphics[width=\methcolimw,height=\methcolimw,keepaspectratio, trim=#3,clip]{#1/#2_images_render_wbg_0.png}%
\includegraphics[width=\methcolimw,height=\methcolimw,keepaspectratio, trim=#3,clip]{#1/#2_shaded_depths_render_medium_wbg_0.png}%
}%
\newcommand{\gtim}[2]{\includegraphics[width=\methcolimw,height=\methcolimw,keepaspectratio]{input_batches/#1_0.png}}
\newcommand{\examplerow}[2]{%
\gtim{#1}{#2}&%
\methodcol{pointclouds_autodec_opacities}{#1}{#2}&%
\methodcol{srn_autodec_opacity}{#1}{#2}&%
\methodcol{nv_autodec}{#1}{#2}&%
\methodcol{idr}{#1}{#2}&%
\methodcol{nerf_opacity}{#1}{#2}&%
\methodcol{transformer_nerf_opacity_dimdown}{#1}{#2}%
}%
\setlength\tabcolsep{0cm}%
\renewcommand{\arraystretch}{0.0}
\centering\small%
\begin{tabular}{cccccccccccc}
Tgt. img.&
\method{IPC} & 
\method{SRN} \cite{sitzmann19scene} & 
\method{NV} \cite{lombardi2019neural} & 
\method{IDR} \cite{yariv2020multiview} & 
\method{NeRF} \cite{mildenhall2020nerf} & 
\textbf{\method{NerFormer}}\vspace{0.1cm}\\%
\examplerow{\detokenize{val_50_2928_8645_13_49_0_56_38_25_98_85_35_34}}{2.5cm 1cm 5cm 3.5cm}\\%
\examplerow{\detokenize{val_167_18184_34441_13_62_63_70_19_82_52_44_34_37}}{0 1.6cm 15cm 3cm}\\%
\examplerow{\detokenize{val_34_1479_4753_16_21_65_58_37_49_0_100_29_15}}{2cm 3.7cm 10cm 3.5cm}\\%
\examplerow{\detokenize{val_374_42274_84517_22_69_15_59_70_20_72_35_5_21}}{0cm 4.8cm 5cm 3.cm}%
\end{tabular}
\let\methcolimw\undefined%
\let\methcolimh\undefined%
\let\methodcol\undefined%
\let\gtim\undefined%
\let\examplerow\undefined%
\caption{%
\textbf{%
Single-scene new-view synthesis on \TheDataset
} depicting a target image from the training video (left), and corresponding synthesized view generated by 
\method{IPC},
\method{SRN} \cite{sitzmann19scene}, 
\method{NV} \cite{lombardi2019neural},
\method{IDR} \cite{yariv2020multiview}, 
\method{NeRF} \cite{mildenhall2020nerf}, 
and our \method{\TheMethod}.%
\label{f:qualitative_sscene}%
}%
\vspace{-0.2cm}%
\end{figure*}

Our main technical contribution aims to alleviate these issues and follows a combination of two main design guidelines:
1) We replace the $f'_\text{MLP}$ with a more powerful architecture \emph{capable of spatial reasoning}.
2) Instead of engineering the WCE aggregation function, we propose to \emph{learn} it.

\paragraph{\TheMethod}
\newcommand{\mlp}{\text{MLP}}%
\newcommand{\mha}{\text{MHA}}%
\newcommand{\te}{\text{TE}}%
\newcommand{\LN}{\text{LN}}%
As a solution, we propose to leverage the popular Transformer architecture \cite{vaswani2017attention}.
Employing a sequence-to-sequence model is intuitive since the set of all WCE embeddings along a projection ray is in fact a depth-ordered descriptor sequence along the ray dimension, and an unordered sequence along the feature pooling dimension.

Formally, 
given a ray $\br_\bu$, 
we define $Z^{\br_\bu} \in \mathbb{R}^{N_\text{S} \times N_\text{src} \times D_\bz}$
as a stacking of un-aggregated WCEs of all ray-points $\bx_j^{\br_u}$:
\begin{equation}
Z^{\br_\bu} = \Big(
\big\{ \bz_\text{WCE}(\bx_j^{\br_u}, I^\text{src}_i, P^\text{src}_i ) \big\}_{i=1}^{n_\text{src}}
\Big)_{j=1}^{N_\text{S}}.
\end{equation}
The \emph{\TheMethod} module $f'_\text{TR}(Z^{\br_\bu}) = f_\text{TR}^\text{HEAD} \circ \te_L \circ \dots \circ \te_1(Z^{\br_\bu})$
replaces the feature backbone $f'_\text{MLP}$ (\cref{s:method_shape_representations}) with a series of $L$ 3D transformer modules $\te_l$ terminated by a weighted pooling head $f_\text{TR}^\text{HEAD}$ (\cref{f:nerformer}).
Here, each 3D transformer module $\te_l$ is a pairing of Transformer Encoder \cite{vaswani2017attention} layers $\te^0(Z)$ and $\te^1(Z)$:
\newcommand{\ZR}{Z_l}%
\begin{align}%
\mha^d_l(\ZR) &= Z_l' = \LN(\mha_l(\ZR, \text{dim=}d) + \ZR)  \\
\te^d_l(\ZR) &= \LN(\mlp_l(Z_l') + Z_l') \\
\te_l(\ZR) &= \te^0_l(\te^1_l(\ZR)) = Z_{l+1},
\end{align}%
$\mha(Z, \text{dim=}d)$ 
is a multi-head attention layer \cite{vaswani2017attention} whose attention vectors span the $d$-th dimension of the input tensor $Z$,
$\mlp$ is a two-layer MLP with ReLU activation,
and $\LN$ is Layer Normalization \cite{ba2016layer}.
Intuitively, the alternation between ray and pooling attention of $\te^0(Z)$ and $\te^1(Z)$ facilitates learning to jointly aggregate WCE features from the source views and ray-march over them respectively.

Finally, $f'_\text{TR}$ is terminated by a weighted pooling head $f_\text{TR}^\text{HEAD}(Z_L)$ that aggregates the second dimension of $Z_L$ output by the final $L$-th 3D transformer module $\te_L$:
$
f_\text{TR}^\text{HEAD}(Z_L) = \sum_{i=1}^{n_\text{src}} 
\omega_i(Z_L) Z_L[:,i,:]
\in \mathbb{R}^{N_\text{S} \times D_\bz},
$
where the weights $\omega_i \in [0,1], \sum_i \omega_i = 1$ 
are output by a linear layer with softmax activation.
We show $\omega_i$ in \cref{f:nerformer_attention}.

The final opacity and coloring functions of \TheMethod\ are thus 
$f_o = f_\text{HEAD} \circ f'_\text{TR}$, 
$c_o = c_\text{HEAD} \circ f'_\text{TR}$ 
respectively, which are rendered with the EA raymarcher (\cref{s:rendering}).

\paragraph{Technical details}%
Training minimizes, with Adam (learning rate $5 \cdot 10^{-4}$), a sum of the RGB MSE error
$\|I^\text{tgt} - \hat I^\text{tgt}\|^2$
and the binary cross entropy between the rendered alpha mask $\hat M^\text{tgt}$ and the ground truth mask $M^\text{tgt}$.
We iterate over batches comprising a randomly sampled target view and 1 to 9 source views of a training video until convergence. 
\section{Experiments}\label{s:exp}

\begin{table}[t]%
\newcommand{\ovrcolw}{0.27cm}%
\scriptsize\centering%
\begin{tabular}{%
@{}
m{1.2cm}
m{\ovrcolw}
m{\ovrcolw}
m{\ovrcolw}
m{0.21cm}
m{1.1cm}
m{\ovrcolw}
m{\ovrcolw}
m{\ovrcolw}
m{\ovrcolw}
m{\ovrcolw}
@{}
}\toprule%
method                             & PSNR        & LPIPS     & $\ell_1^\text{depth}$ & IoU       & 
method & PSNR        & LPIPS     & $\ell_1^\text{depth}$ & IoU  \\ \midrule
\transformerAnerfAopacityAdimdown  & \b{23.3}  & \b{0.17}  & \b{0.40}  & \bb{0.96} & \srnAautodecAopacity               & 20.4 & 0.21 & 0.60 & 0.93  \\
\nerfAviewpoolAopacity             & 21.0      & 0.19      & 0.74      & 0.91      & \srnAharmonicAviewpoolAopacity     & 16.9 & 0.30 & 0.60 & 0.75  \\
\nerfAopacity                      & \bb{23.6} & \b{0.17}  & \bb{0.38} & \b{0.95}  & \srnAviewpoolAopacity              & 15.8 & 0.26 & 0.64 & 0.80  \\
\nvAautodec                        & 22.2      & 0.20      & 0.91      & 0.91      & \srnAharmonicAautodecAopacity      & 16.9 & 0.30 & 0.59 & 0.75  \\
\nvAviewpool                       & 18.7      & 0.25      & 0.85      & 0.90      & \dvrAautodec                       & 15.0 & 0.33      & 0.89                  & 0.68  \\
\idr                               & 18.5      & \bb{0.15} & 0.81      & 0.92      & \dvrAharmonicAautodec              & 15.8 & 0.38      & 0.74                  & 0.65  \\
\pointcloudsAviewpoolAopacities    & 13.9      & 0.25      & 1.59      & 0.83      & \softrasAautodec                   & 17.3 & 0.22      & 0.69                  & 0.91  \\
\pointcloudsAautodecAopacities     & 13.8      & 0.25      & 1.58      & 0.83      & \\ \bottomrule
\end{tabular}
\vspace{-0.2cm}%
\caption{%
\textbf{Single-scene new-view synthesis results on \TheDataset}
comparing the baseline approaches \cite{mildenhall2020nerf,lombardi2019neural,yariv2020multiview,sitzmann19scene,niemeyer2020differentiable,wang2018pixel2mesh}, IPC, their variants with Warp-conditioned Embeddings (\method{+\textbf{WCE}}) or Positional Embedding (+$\boldmath{\gamma}$), and our \TheMethod\ (the \bb{best} / \b{2nd best} result).%
\vspace{-0.2cm}%
}\label{t:results_sscene}%
\end{table}

\begin{table*}[t]%
\setlength\tabcolsep{0.05cm}
\scriptsize\centering
\newcommand{\isad}{\textsuperscript{\textdagger}}
\begin{tabular}{
@{}
l@{\extracolsep{0.1cm}}
rrrr@{\extracolsep{0.1cm}}
rrrr@{\extracolsep{0.1cm}}
rrrrr@{\extracolsep{0.1cm}}
rrrrr@{\extracolsep{0.1cm}}
rrrrrr
@{}
}

               & \multicolumn{8}{c}{\textbf{(a) Average statistics}} 
               & \multicolumn{10}{c}{\textbf{(b) PSNR @ \# source views}} 
               & \multicolumn{6}{c}{\textbf{(c) PSNR @ target view difficulty}} \\ 
               \cmidrule{2-9} \cmidrule{10-19} \cmidrule{20-25}

& \multicolumn{4}{c}{\traintestset} 
& \multicolumn{4}{c}{\testtestset} 
& \multicolumn{5}{c}{\traintestset}
& \multicolumn{5}{c}{\testtestset} 
& \multicolumn{3}{c}{\traintestset} 
& \multicolumn{3}{c}{\testtestset} 
\\

                Method                         & PSNR & LPIPS & $\ell_1^\text{depth}$ & IoU  & PSNR & LPIPS & $\ell_1^\text{depth}$  & IoU       & 9    & 7 & 5 & 3 & 1 & 9  & 7 & 5 & 3  & 1 & easy     & med.    & hard    & easy     & med.   & hard    \\ \cmidrule{1-1} \cmidrule{2-5} \cmidrule{6-9} \cmidrule{10-14} \cmidrule{15-19} \cmidrule{20-22} \cmidrule{23-25}
\transformerAnerfAopacityAdimdown              & \bb{17.9} & 0.26      & 0.87        & 0.82      & \bb{17.6} & \b{0.27}  & 0.91      & \b{0.81}  & \bb{19.3} & \bb{19.0} & \bb{18.3} & \b{17.4}  & 15.6      & \bb{18.9} & \bb{18.6} & \bb{18.1} & \bb{17.1} & \bb{15.1} & \bb{18.9} & 15.5      & \b{14.6}  & \bb{18.6} & \bb{14.9} & 14.7 \\
\srnAharmonicAviewpoolAopacity                 & \b{17.6}  & \b{0.24}  & \bb{0.28}   & \b{0.89}  & 14.4      & \b{0.27}  & \b{0.40}  & \b{0.81}  & \b{18.0}  & \b{18.0}  & \b{17.8}  & \bb{17.6} & \bb{16.8} & 14.6      & 14.5      & 14.6      & 14.5      & \b{13.9}  & \b{18.0}  & \bb{16.8} & \bb{16.0} & 14.7      & 13.6      & \bb{15.1} \\
\srnAviewpoolAopacity                          & 16.6      & 0.26      & \b{0.31}    & 0.87      & \b{14.6}  & \b{0.27}  & \bb{0.36} & \bb{0.82} & 17.0      & 17.0      & 16.7      & 16.4      & 15.8      & \b{14.9}  & \b{14.8}  & \b{14.8}  & 14.6      & \b{13.9}  & 16.9      & \b{15.7}  & 14.5      & \b{14.9}  & \b{13.7}  & \b{14.8} \\
\nerfAviewpoolAopacity                         & 14.3      & 0.27      & 2.14        & 0.72      & 13.8      & \b{0.27}  & 2.23      & 0.70      & 14.3      & 15.0      & 14.9      & 14.7      & 14.2      & 12.6      & 14.5      & 14.4      & 14.2      & 13.8      & 12.1      & 13.6      & 13.6      & 14.4      & 13.0      & 13.0 \\
\pointcloudsAviewpoolAopacities                & 14.1      & 0.36      & 2.12        & 0.70      & 13.5      & 0.37      & 2.24      & 0.69      & 14.4      & 14.4      & 14.2      & 14.1      & 13.4      & 13.8      & 13.8      & 13.7      & 13.6      & 12.6      & 14.4      & 13.7      & 13.4      & 13.8      & 12.8      & 12.2 \\
\softrasAviewpool                              & 17.2      & \bb{0.23} & 0.50        & \bb{0.91} & 12.4      & \bb{0.26} & 2.49      & 0.69      & 17.6      & 17.5      & 17.4      & 17.1      & \b{16.2}  & 12.6      & 12.5      & 12.5      & 12.5      & 12.1      & 17.5      & \bb{16.8} & \bb{16.0} & 12.6      & 11.8      & 13.2 \\
\nvAviewpool                                   & 12.3      & 0.34      & 2.87        & 0.54      & 11.6      & 0.35      & 3.01      & 0.53      & 12.5      & 12.5      & 12.3      & 12.2      & 12.0      & 11.7      & 11.6      & 11.6      & 11.6      & 11.3      & 12.4      & 12.0      & 13.6      & 11.7      & 11.2      & 12.0 \\\hline
\srnAharmonicAautodecAopacity\method{+AD}      & \bb{21.7} & \bb{0.21} & \b{0.31}    & 0.89      & - & - & - & - & \bb{21.7} & \bb{21.7} & \bb{21.7} & \bb{21.7} & \bb{21.8} & - & - & - & - & - & \bb{21.7} & \bb{21.3} & \bb{19.7} & - & - & - \\
\srnAautodecAopacity\method{+AD}               & \b{21.2}  & \bb{0.21} & \bb{0.23}   & \bb{0.94} & - & - & - & - & \b{21.2}  & \b{21.2}  & \b{21.2}  & \b{21.2}  & \b{21.3}  & - & - & - & - & - & \b{21.2}  & \b{20.8}  & \b{19.2}  & - & - & - \\
\nvAautodec\method{+AD}                        & 19.7      & \b{0.23}  & 0.41        & \b{0.93}  & - & - & - & - & 19.7      & 19.7      & 19.6      & 19.6      & 19.7      & - & - & - & - & - & 19.7      & 19.3      & 17.7      & - & - & - \\
\nerfAautodecAopacity                          & 17.1      & 0.25      & 0.55        & 0.92      & - & - & - & - & 17.1      & 17.0      & 17.0      & 17.0      & 17.1      & - & - & - & - & - & 17.1      & 16.9      & 15.8      & - & - & - \\
\softrasAautodec\method{+AD}                   & 16.1      & 0.24      & 0.74        & 0.89      & - & - & - & - & 16.1      & 16.1      & 16.1      & 16.1      & 16.2      & - & - & - & - & - & 16.1      & 16.0      & 15.1      & - & - & - \\
\pointcloudsAautodecAopacities\method{+AD}     & 14.0      & 0.37      & 2.20        & 0.69      & - & - & - & - & 14.4      & 14.3      & 14.1      & 14.0      & 13.3      & - & - & - & - & - & 14.3      & 13.6      & 13.2      & - & - & - \\
\idrAautodec                                   & 13.7      & 0.25      & 1.74        & 0.88      & - & - & - & - & 13.7      & 13.7      & 13.7      & 13.7      & 13.8      & - & - & - & - & - & 13.7      & 13.7      & 14.0      & - & - & - \\
\dvrAharmonicAautodec\method{+AD}              & 7.6       & 0.29      & 3.24        & 0.09      & - & - & - & - & 7.6       & 7.6       & 7.5       & 7.5       & 7.7       & - & - & - & - & - & 7.6       & 7.7       & 9.0       & - & - & - \\
\dvrAautodec\method{+AD}                       & 7.6       & 0.29      & 3.01        & 0.06      & - & - & - & - & 7.6       & 7.6       & 7.5       & 7.5       & 7.7       & - & - & - & - & - & 7.6       & 7.7       & 9.0       & - & - & -\\ \bottomrule%
\end{tabular}
\vspace{-0.2cm}%
\caption{%
\textbf{Category-centric new-view synthesis results on \TheDataset}
comparing mean performance over 10 object categories.
Reported are: (a) the main metrics averaged over the \traintestset/\testtestset set;
(b) PSNR averaged over test samples with a specific number of source views 
and; (c) the samples within one of three viewpoint difficulty bins.
Methods with \method{+AD} are autodecoders and thus not applicable to the \testtestset set. 
(\bb{best} / \b{2nd best})%
}\label{t:results_mscene}%
\vspace{-0.2cm}%
\end{table*}

\paragraph{Datasets}
In order to evaluate and train reconstruction methods on our dataset, we split the \nvideos collected \TheDatasetAbbrev videos into 4 different sets as follows.
For each of the \ncategories categories, we split its videos to a \trainset and a \testset set in 9:1 ratio.
For each video, we further define a set of frames that are removed from the training set by
randomly dividing each \trainset video in a 8:2 ratio to 80 \traintrainset training and 20 \traintestset holdout frames.
\testset video frames are split according to the same protocol resulting in \testtrainset and \testtestset sets.
As all baselines require a bounded scene coordinate frame, we zero-center each point cloud and normalize the scale by dividing with an average over per-axis standard deviations.

\paragraph{Evaluation protocol}\label{s:metrics}
All evaluations focus on new RGB+D-view synthesis where a method,
given a known target camera $P^\text{tgt}$ and a set of source views $\{I^\text{src}_i\}_{i=1}^{n_\text{src}}$, renders new target RGB view $\hat I^\text{tgt}$ and depth $\hat D^\text{tgt}$.
Four metrics are reported: The peak-signal-to-noise ratio (PSNR) and the LPIPS distance \cite{zhang2018perceptual} between $\hat I^\text{tgt}$ and the ground truth target RGB frame $I^\text{tgt}$, the Jaccard index between the rendered/g.t. object mask $\hat M^\text{tgt}$/$M^\text{tgt}$ and the $\ell_1$ distance $\ell_1^\text{depth}$ between the rendered depth $\hat D^\text{tgt}$ and the ground truth point cloud depth render $D^\text{tgt}$.
Note that PSNR and $\ell_1^\text{depth}$ are only aggregated over the foreground pixels 
$\{\bu_\text{fg} | M^\text{tgt}[\bu_\text{fg}] = 1 \}$.
Metrics are evaluated at an image resolution of 800x800 and 400x400 pixels for single-scene (\cref{s:exp_single_scene}) and category-specific reconstruction (\cref{s:exp_multi_scene}) respectively.

\subsection{Evaluated methods}%
We implemented a total of 15 approaches that leverage the majority of common shape representation types.

\paragraph{Implicit surfaces}%
Among implicit surface estimators we selected the seminal opacity-based \method{NeRF}; and \method{IDR} \cite{yariv2020multiview} and \method{DVR} \cite{niemeyer2020differentiable} expressing shapes as a SDF $f_d$ and render with sphere-tracing.
We further benchmark \method{SRN} \cite{sitzmann19scene} which implements an implicit learned LSTM renderer.
Importantly, following \cref{s:method_boosting_positional,s:method_boosting_wce}, 
we evaluate the modifications \method{SRN-$\gamma$}, \method{DVR-$\gamma$} that endow SRN and DVR respectively with positional embedding $\gamma$.
Furthermore, \method{SRN-$\gamma$-WCE}, \method{SRN-WCE}, \method{NeRF-WCE} \cite{henzler2021unsupervised} complement SRN and NeRF with the Warp-conditioned Embedding \cite{henzler2021unsupervised}.
Finally, we also evaluate our \TheMethod  method.

\paragraph{Voxel grids}
Neural Volumes (\method{NV}) \cite{lombardi2019neural} represents the State of the Art among voxel grid predictors.
Similar to implicit methods, we also combine \method{NV} with WCE.

\paragraph{Point clouds} 
In order to compare with a point cloud-based method, we devised an Implicit Point Cloud (\method{IPC}) baseline which represents shapes with a colored set of 3D points, converts the set into an implicit surface and then renders it with the EA raymarcher.
We note that \method{IPC} is strongly inspired by SynSin \cite{ravi2020pytorch3d,wiles2020synsin} (see supplementary).

\paragraph{Meshes} We benchmark \method{P3DMesh} - the best-performing variant of PyTorch3D's soft mesh rasterizer from \cite{ravi2020pytorch3d} inspired by Pixel2Mesh \cite{wang2018pixel2mesh}, which deforms an initial spherical mesh template with a fixed topology with a series of convolutions on the mesh graph.

\begin{figure*}[th]
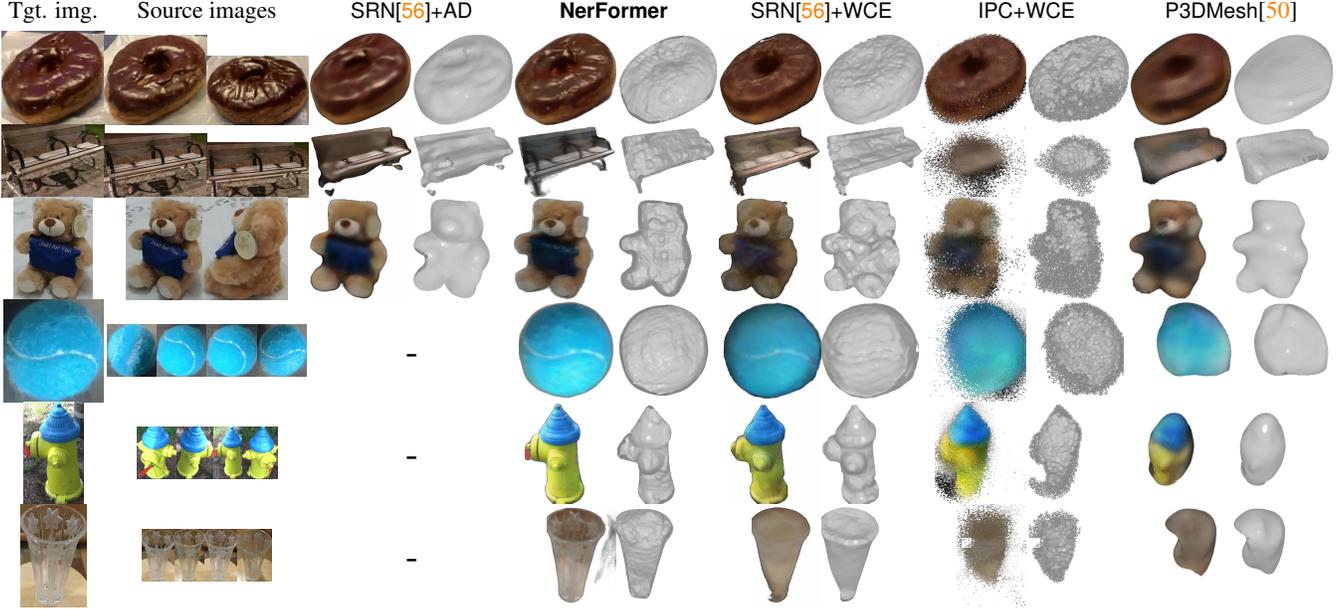

\centering%
\graphicspath{{figures/qual_multiscene/}}
\newcommand{\methcolimw}{0.078\linewidth}
\newcommand{\methodcol}[3]{%
\includegraphics[width=\methcolimw,height=\methcolimw,keepaspectratio, trim=#3,clip]{#1/#2_images_render_wbg_0.png}%
\includegraphics[width=\methcolimw,height=\methcolimw,keepaspectratio, trim=#3,clip]{#1/#2_shaded_depths_render_medium_wbg_0.png}%
}%
\newcommand{\gtim}[2]{\includegraphics[width=\methcolimw,height=\methcolimw,keepaspectratio]{input_batches/#1_#2.png}}
\newcommand{\examplerow}[2]{%
\gtim{#1}{0}&%
\gtim{#1}{1}%
\gtim{#1}{2}%
&%
\methodcol{srn_ad}{#1}{#2}&%
\methodcol{nerformer}{#1}{#2}&%
\methodcol{srn}{#1}{#2}&%
\methodcol{pcl}{#1}{#2}&%
\methodcol{softras}{#1}{#2}%
}
\newcommand{\examplerowtest}[2]{%
\gtim{#1}{0}&%
{%
\raisebox{.5\height}{%
\scalebox{0.5}{%
\gtim{#1}{3}%
\gtim{#1}{4}%
\gtim{#1}{1}%
\gtim{#1}{2}%
}%
}%
}&%
\raisebox{0.5cm}{%
\Large{-}%
}%
&%
\methodcol{nerformer}{#1}{#2}&%
\methodcol{srn}{#1}{#2}&%
\methodcol{pcl}{#1}{#2}&%
\methodcol{softras}{#1}{#2}%
}
\setlength\tabcolsep{0cm}%
\renewcommand{\arraystretch}{0.0}
\centering\small%
\begin{tabular}{ccccccccc}
Tgt. img.&%
Source images&%
\method{SRN\cite{sitzmann19scene}+AD}&
\textbf{\method{\TheMethod}} &
\method{SRN\cite{sitzmann19scene}+WCE}& 
\method{IPC+WCE} & 
\softrasAviewpool\cite{ravi2020pytorch3d}%
 \vspace{0.1cm} \\
\examplerow{\detokenize{train_339_35131_63274_43_88_20_44_93_67_85_4}}{1.5cm 2.7cm 1.5cm 1.5cm}\\
\examplerow{\detokenize{train_106_12699_27006_91_95_97_62_20_25_90_101_60_51}}{1.2cm 5.5cm 1.2cm 0.5cm}\\
\examplerow{\detokenize{train_246_26303_51369_8_12_36_21_14_69}}{1.2cm 1.5cm 1.8cm 1.5cm}\\
\examplerowtest{\detokenize{val_141_16155_30795_68_75_65_96}}{1.5cm 1.2cm 1.5cm 1.2cm}\\
\examplerowtest{\detokenize{val_134_15451_31119_7_85_59_70_35_96_97_50}}{1.2cm 1.5cm 3cm 1.5cm}\\
\examplerowtest{\detokenize{val_196_21116_42822_27_26_50_29_12_52_35_64_22_2}}{1.2cm 1.2cm 5cm 1.2cm}\\
\end{tabular}
\vspace{-0.2cm}%
\caption{%
\textbf{%
Category-centric 3D reconstruction on \TheDataset
} depicting a target image, 
known source images $\{ I^\text{src}_i\}$,
and a synthesized new view.
The first/last 3 rows are from the \trainset/\testset set (\method{SRN+AD} is not applicable to \testset).%
\label{f:qualitative_mscene}%
}%
\vspace{-0.28cm}%
\end{figure*}

\subsection{Single-scene reconstruction}\label{s:exp_single_scene}
We first task the approaches to independently reconstruct individual object videos.
More specifically, given a \testset video, every baseline is trained to reproduce the video's \testtrainset frames
(by minimizing method-specific loss functions such as $\ell_2$ RGB error), and evaluated by comparing the renders from the given \testtrainset camera viewpoints to the corresponding ground truth images/depths/masks.
Since training on all $\sim$2k \testset videos is prohibitively expensive (each baseline trains at least for 24 hours on a single GPU), we test on 40 randomly selected \testset videos (two from each of 20 random classes). 
Quantitative / qualitative results are in \cref{f:qualitative_sscene} / \Cref{t:results_sscene}.

\method{\TheMethod} is either the best or the second best across all metrics.
\method{NeRF+WCE} is beaten by vanilla \method{NeRF} which suggests that the noisy WCE embeddings can hurt performance without \method{\TheMethod}'s spatial reasoning.
Interestingly, \method{IDR}'s realistic, but less detailed renders win in terms of LPIPS, but are inferior in PSNR. Furthermore, we observed a large discrepancy between the train and test PSNR of \method{$\gamma$/WCE}-endowed \method{SRN}.
This shows that, for the single-scene setting, increasing the model expressivity with WCE or $\gamma$ can lead to overfitting to the training views.

\subsection{Learning 3D Object Categories}\label{s:exp_multi_scene}
Our main task is learning category-centric 3D models.
In more detail, a single model is trained on all \traintrainset frames of an object category and evaluated on 1000 randomly generated test samples from \traintestset and \testtestset sets of the category.
Each train or test sample is composed of a single target image $I^\text{tgt}$ and randomly selected source views $\{I^\text{src}\}_i^{n_\text{src}}$, where $n_\text{src}$ is randomly picked from $\{1, 3, 5, 7, 9\}$.
Since training a model for each of the 50 object categories is prohibitively expensive (each method takes at least 7 days to train), we chose a subset of 10 categories for evaluation.

\paragraph{Baselines}%
For each method we evaluate the ability to represent the shape space of training object instances by turning it into an \emph{autodecoder} \cite{bojanowski2017optimizing} which, in an encoder-less manner, learns a separate latent embedding $\bz_\text{scene}(\text{sequence}_\text{ID})$ for each \trainset scene as a free training parameter.
Since autodecoders (abbreviated with the \textbf{\method{+AD}} suffix) only represent the \trainset set, 
we further compare to all WCE-based methods which can additionally reconstruct \testset videos.
Note that, as remarked in \cite{henzler2021unsupervised} and \cref{s:method_boosting_wce}, the alternative encoding $\bz_\text{global} = \Phi_\text{CNN}(I^\text{src})$ is deemed to fail due to the world coordinate ambiguity of the training SfM reconstructions.

\paragraph{Results}%
Quantitative and qualitative comparisons are shown in \cref{t:results_mscene,f:qualitative_mscene} respectively.
Besides average metrics over both test sets, we also analyze the dependence on the number of available source views, and on the difficulty of the target view.
For the latter, test frames are annotated with a measure of their distance to the set of available source views, and then split into 3 different difficulty bins. Average per-bin PSNR is reported.
The supplementary details the distance metric and difficulty bins.

While \method{SRN+AD} has the best performance across all metrics on \traintestset; on the \testtestset set, where \method{SRN+AD} is inapplicable, our \method{\TheMethod} is the best for most color metrics.
Among implicit methods, the SDF-based \method{DVR} and \method{IDR} are outperformed by the opacity-based \method{NeRF}, with both \method{DVR+WCE} and \method{IDR+WCE} failing to converge.
This is likely because regressing SDFs is more challenging than classifying 3D space with binary labels opaque/transparent.
Finally, we observed poor performance of \softrasAviewpool, probably due to the inability of meshes to represent complicated real geometries and textures.

\section{Conclusion}\label{s:ccl}
We have introduced \TheDataset (\TheDatasetAbbrev), a dataset of in-the-wild object-centric videos capturing 50 object categories with camera and point cloud annotations.

We further contributed \TheMethod which is a marriage between Transformer and neural implicit rendering that can reconstruct 3D object categories from \TheDatasetAbbrev with better accuracy than a total of 14 other tested baselines.

The \TheDatasetAbbrev collection effort still continues at a steady pace of $\sim$500 videos per week which we plan to release in the near future.

\bibliographystyle{ieee_fullname}%

\ifx\arxivsubmission\undefined

{\small%
\bibliography{refs,vedaldi_general,vedaldi_specific}
}
\clearpage
\appendix
\setcounter{figure}{0} \renewcommand{\thefigure}{\Roman{figure}}
\setcounter{table}{0} \renewcommand{\thetable}{\Roman{table}}

\begin{strip}%
 \centering
 \Large
 \textbf{%
 \TheDataset:\\
{\Large Large-Scale Learning and Evaluation of Real-life 3D Category Reconstruction}\\
 \vspace{0.3cm} \textit{Supplementary material}
 }\\
\vspace{1cm}
\includegraphics[width=\linewidth]{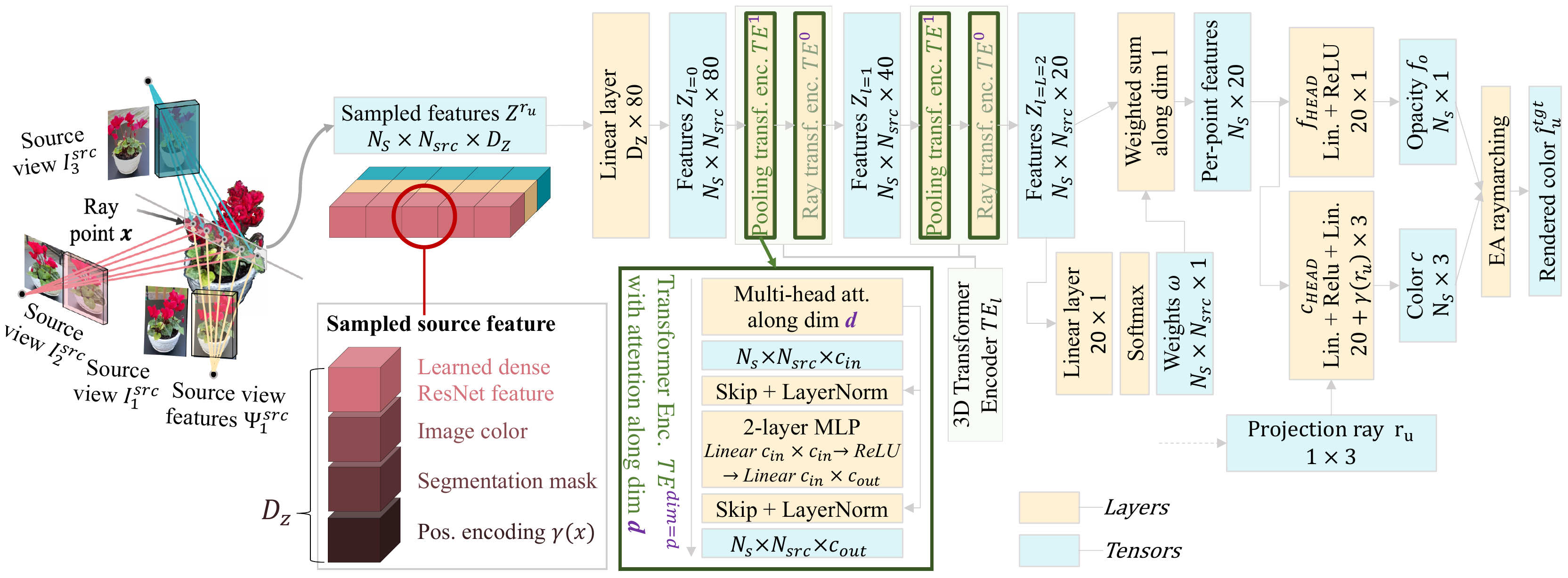}
\captionof{figure}{A detailed illustration of the architecture of \method{\TheMethod}.\label{f:sup_nerformer}}
\end{strip}

\begin{table*}[th]%
\setlength\tabcolsep{0.05cm}
\scriptsize\centering
\newcommand{\isad}{\textsuperscript{\textdagger}}
\begin{tabular}{
@{}
l@{\extracolsep{0.1cm}}
rrrr@{\extracolsep{0.1cm}}
rrrr@{\extracolsep{0.1cm}}
rrrrr@{\extracolsep{0.1cm}}
rrrrr@{\extracolsep{0.1cm}}
rrrrrr
@{}
}
               & \multicolumn{8}{c}{\textbf{(a) Average statistics}} 
               & \multicolumn{10}{c}{\textbf{(b) PSNR @ \# source views}} 
               & \multicolumn{6}{c}{\textbf{(c) PSNR @ target view difficulty}} \\ 
               \cmidrule{2-9} \cmidrule{10-19} \cmidrule{20-25}
& \multicolumn{4}{c}{\traintestset} 
& \multicolumn{4}{c}{\testtestset} 
& \multicolumn{5}{c}{\traintestset}
& \multicolumn{5}{c}{\testtestset} 
& \multicolumn{3}{c}{\traintestset} 
& \multicolumn{3}{c}{\testtestset} 
\\ 
                Method            & PSNR & LPIPS & $\ell_1^\text{depth}$ & IoU   & PSNR & LPIPS & $\ell_1^\text{depth}$ & IoU         & 9         & 7           & 5           & 3           & 1           & 9         & 7         & 5         & 3         & 1         & easy      & med.      & hard    & easy     & med.   & hard \\ \cmidrule{1-1} \cmidrule{2-5} \cmidrule{6-9} \cmidrule{10-14} \cmidrule{15-19} \cmidrule{20-22} \cmidrule{23-25}
\transformerAnerfAopacityAdimdown & \b{16.5}  & \bb{0.24} & 3.67      & \b{0.76}  & \bb{15.7} & \bb{0.24} & \b{1.82}  & \b{0.75}  & \bb{17.5} & \b{17.3}  & \b{16.9}  & \b{16.3}  & 14.8      & \bb{16.7} & \bb{16.4} & \bb{16.1} & \bb{15.5} & \bb{13.9} & \b{17.3}  & 14.7      & 12.8      & \bb{16.5} & \bb{13.7} & \b{11.2} \\
\srnAviewpoolAopacity             & 16.3      & \b{0.25}  & \b{0.37}  & \bb{0.81} & \b{14.2}  & \b{0.27}  & \bb{0.47} & \bb{0.77} & 16.6      & 16.6      & 16.5      & 16.2      & \b{15.6}  & \b{14.4}  & \b{14.3}  & \b{14.3}  & \b{14.2}  & \b{13.5}  & 16.6      & \b{15.5}  & \b{12.9}  & \b{14.4}  & \b{13.4}  & \bb{11.4} \\
\srnAharmonicAviewpoolAopacity    & \bb{17.1} & 0.25      & \bb{0.35} & \bb{0.81} & 13.7      & 0.28      & \bb{0.47} & 0.73      & \b{17.4}  & \bb{17.4} & \bb{17.3} & \bb{17.0} & \bb{16.3} & 14.0      & 13.8      & 13.9      & 13.7      & 13.2      & \bb{17.4} & \bb{16.3} & \bb{14.4} & 14.0      & 13.1      & 10.6 \\
\nerfAviewpoolAopacity            & 12.6      & 0.27      & 6.21      & 0.54      & 11.6      & \b{0.27}  & 4.54      & 0.51      & 13.0      & 13.0      & 12.8      & 12.6      & 11.6      & 11.9      & 11.8      & 11.8      & 11.6      & 10.8      & 12.9      & 12.1      & 9.4       & 11.9      & 11.1      & 8.7  \\ \bottomrule%
\end{tabular}
\caption{%
\textbf{Results on all 50 classes from \TheDatasetAbbrev}
comparing the 4 best-performing methods from \cref{t:results_mscene}.%
\label{t:results_mscene_full}}%
\end{table*}

In what follows, we provide additional quantitative results (\cref{s:sup_results}), technical details of \method{\TheMethod} and the baselines (\cref{s:sup_tech_details}), and details of the Human-in-the-loop 3D annotation process (\cref{s:sup_human_il}).

\section{Additional results} \label{s:sup_results}

\subsection{Results on all 50 categories}
While \cref{t:results_mscene} in the main paper provides results on a subset of 10 object categories for all baselines, for completeness, in \cref{t:results_mscene_full}, we provide evaluation on all 50 object classes for 4 best-performing baselines according to results reported in \cref{t:results_mscene}: \method{\TheMethod, SRN+WCE, SRN+$\gamma$+WCE}, and \method{NeRF+WCE}.

Similar to \cref{t:results_mscene}, on the \testtestset set, \method{\TheMethod} is the best in all color-based metrics, suggesting that \method{SRN} and \method{NeRF+WCE} are prone to overfitting to the training scenes.
While SRN outperforms \method{\TheMethod} in some cases on \traintestset, we note that the autodecoders would likely yield superior performance on \traintestset due to their ability to capture the information from all views of a training scene in the latent scene-specific encoding.

\subsection{Convergence speed}
\Cref{f:convergence_sscene} further analyzes training convergence on the single-scene new-view synthesis task.
For each method and epoch, we plot the average and standard deviation over of the per-epoch mean PSNRs of each of the 40 test scenes.
The fastest converging methods are \method{SRN}, \method{IDR}, \method{\TheMethod}, \method{NeRF} which also top the performance in \cref{t:results_sscene}, indicating a significant correlation between convergence rate and performance.
Furthermore, there is a large discrepancy between the train and test PSNR of \method{$\gamma$/WCE}-endowed \method{SRN}. 
This shows that, for the single-scene setting, increasing the model expressivity with WCE or $\gamma$ can lead to overfitting to the training views.

\subsection{Execution speed}
\Cref{t:results_speed} contains an evaluation of execution times for all methods from \cref{t:results_sscene}.
Here, each row reports an average time to render an 800x800 pixel image on NVIDIA Tesla V100 GPU.

\begin{table}[h]
\newcommand{\speedcolw}{0.27cm}%
\footnotesize\centering%
\begin{tabular}{%
@{}
lrlr
@{}
}\toprule%
method                              & time {[}sec{]} & method                           & time {[}sec{]} \\ \midrule
\transformerAnerfAopacityAdimdown & 178.41          & \srnAautodecAopacity            & 1.00           \\
\nerfAviewpoolAopacity             & 113.82         & \srnAharmonicAautodecAopacity  & 1.19           \\
\nerfAopacity                       & 23.82          & \srnAharmonicAviewpoolAopacity & 4.20           \\
\nvAautodec                         & 0.37           & \srnAviewpoolAopacity           & 5.34           \\
\nvAviewpool                        & 0.41           & \dvrAautodec                     & 196.94         \\
\idr                                 & 69.11          & \dvrAharmonicAautodec           & 204.39         \\
\pointcloudsAautodecAopacities     & 0.15           & \softrasAautodec                 & 0.09           \\
\pointcloudsAviewpoolAopacities    & 0.16           &                                  &               \\ \bottomrule
\end{tabular}%
\caption{%
\textbf{Average rendering time} of an 800x800 pixel image comparing all methods from \cref{t:results_sscene}.%
\label{t:results_speed}%
}%
\end{table}

\subsection{Test-time autodecoder optimization}
In \cref{t:results_mscene_ad_test}, we provide an extension of \cref{t:results_mscene} containing the evaluation of the best-performing autodecoding methods at test-time.
First, each method is first trained on \traintrainset. 
Then, during evaluation, the latter freezes the trained weights and optimizes the input latent code for a given set of source frames from a test sequence.
The latent codes are optimized with Adam until convergence, decaying the learning rate 10-fold whenever the optimization objective plateaus.

We observed that the latent code optimization was mostly failing for \method{NeRF} and \method{NV}.
On the other hand, \method{SRN} gave slightly better performance, which we attribute to the higher smoothness of the implicit function compared to the \method{NeRF} and \method{NV} (\method{SRN} contains normalization layers while the other baselines are bare MLPs interleaving linear layers and ReLUs).

\begin{figure}[t]%
\centering%
\includegraphics[%
width=\linewidth,%
trim=0.3cm 0.5cm 0.3cm 0.45cm,%
clip,%
]{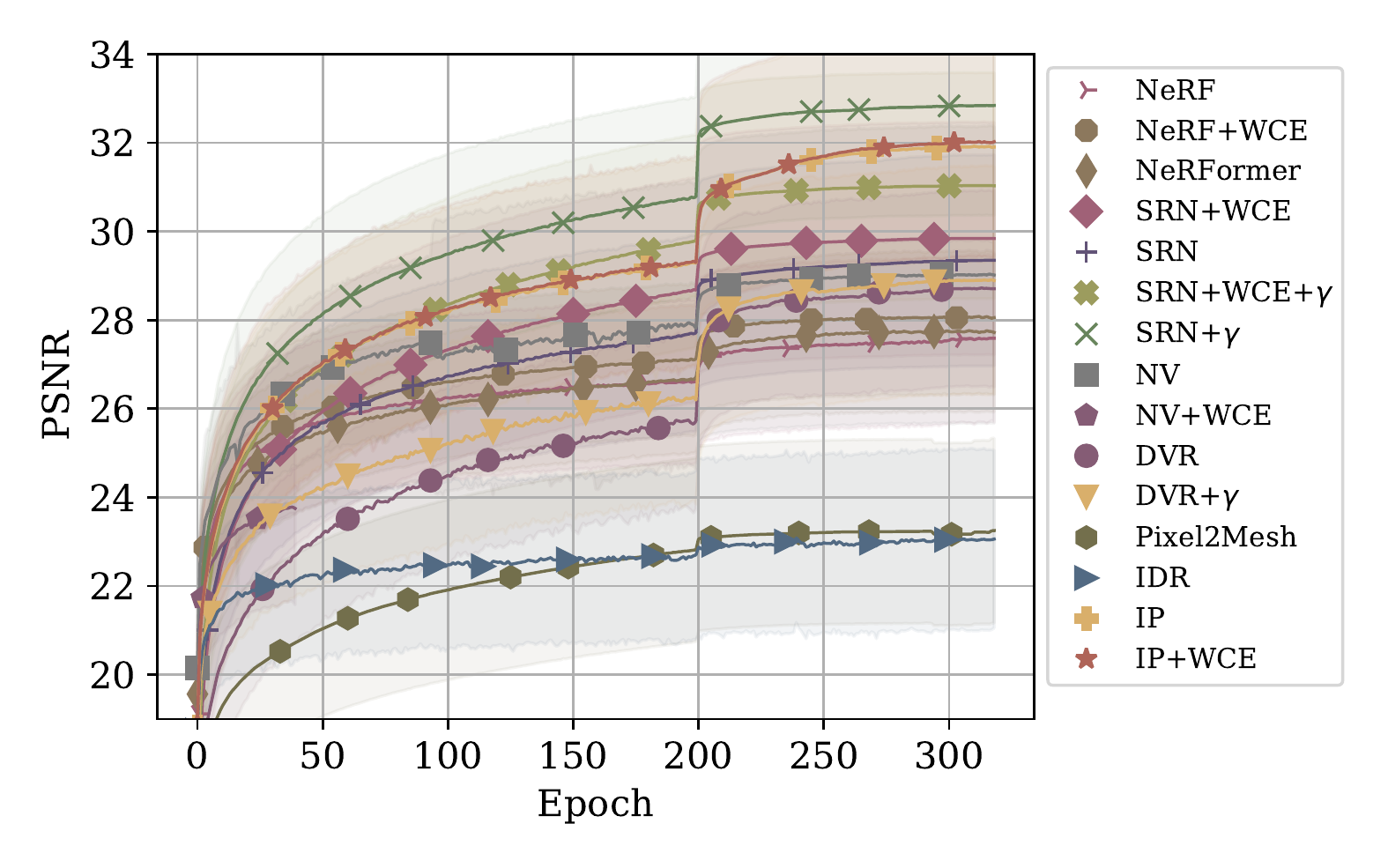}%
\vspace{-0.2cm}%
\caption{%
\textbf{Convergence speed on single-scene new-view-synthesis} showing the mean and std. dev. over the per-scene training PSNRs for each method.%
\label{f:convergence_sscene}%
}%
\end{figure}

\subsection{Estimating new-view difficulty}
\Cref{t:results_mscene} contains metrics evaluated separately for three viewpoint difficulty bins.
Here, we detail the process of estimating the difficulty of a testing target view.

\paragraph{Camera difficulty $\mathcal{D}$} Given a target camera $P^\text{tgt}$ and a set of available source views $\{P^\text{src}_i\}_{i=1}^{N_\text{src}}$,
the difficulty of the target view $\mathcal{D}(P^\text{tgt}) \in [0, 1]$ is quantified as the average of the two lowest distances $d(P^\text{tgt}, P^\text{src}_i)$ between the target view and each of the source views.

\begin{table*}[th]
\setlength\tabcolsep{0.05cm}
\footnotesize\centering
\begin{tabular}{
@{}
l@{\extracolsep{0.3cm}}
rrrr@{\extracolsep{0.3cm}}
rrrrr@{\extracolsep{0.3cm}}
rrr
@{}
}\toprule
& \multicolumn{4}{c}{\textbf{(a) Average statistics}} 
& \multicolumn{5}{c}{\textbf{(b) PSNR @ \# src. views}} 
& \multicolumn{3}{c}{\textbf{(c) PSNR @ tgt. difficulty}} \\ 
\cmidrule{2-5} \cmidrule{6-10} \cmidrule{11-13}
& \multicolumn{4}{c}{\testtestset} 
& \multicolumn{5}{c}{\testtestset} 
& \multicolumn{3}{c}{\testtestset} 
\\               
Method                         &  PSNR & LPIPS & $\ell_1^\text{depth}$  & IoU      & 9  & 7 & 5 & 3  & 1 & easy     & medium   & hard    \\ \cmidrule{1-1} \cmidrule{2-5} \cmidrule{6-10} \cmidrule{11-13}
\transformerAnerfAopacityAdimdown                & \bb{17.6}              & \bb{0.27}           & 0.91                         & \bb{0.81}         & \bb{18.9}              & \bb{18.6}              & \bb{18.1}              & \bb{17.1}              & \bb{15.1}              & \bb{18.6}               & \bb{14.9}                 & \bb{14.7}               \\
\srnAharmonicAautodecAopacity\method{+AD}        & \b{13.2}               & 0.29                & \b{0.48}                     & 0.71              & 13.6                   & 13.5                   & 13.3                   & 13.1                   & 12.4                   & 13.5                    & 11.6                      & \b{11.8}                    \\
\srnAautodecAopacity\method{+AD}                 & 13.8                   & \b{0.28}            & \bb{0.45}                    & \b{0.74}          & \b{14.3}               & \b{14.3}               & \b{14.0}               & \b{13.6}               & \b{12.6}               & \b{14.2}                & \b{12.5}                  & 11.1                    \\
\nvAautodec\method{+AD}                          & 11.4                   & 0.53                & 1.29                         & 0.47              & 11.5                   & 11.2                   & 11.3                   & 11.5                   & 11.5                   & 11.4                    & 11.3                      & 8.0                     \\
\nerfAautodecAopacity                            & 10.6                   & 0.32                & 4.42                         & 0.49              & 10.7                   & 10.5                   & 10.4                   & 10.7                   & 10.4                   & 10.7                    & 10.3                      & 3.6                    \\ \bottomrule
\end{tabular}
\caption{%
\textbf{Autodecoder latent optimization on \testtestset} extending the results in \cref{t:results_mscene}.
Each method labelled with \method{+AD} is first trained on \traintrainset. 
During evaluation the latter fixes the trained weights and optimizes the input latent code for a given set of source frames from a test sequence.
For context, we also compare to \method{\TheMethod}, which is not an autodecoder.
\label{t:results_mscene_ad_test}}
\end{table*}

\paragraph{Camera distance $d^\text{cam}$} The distance $d^\text{cam}(P_i, P_j) \in [0,1]$ between two cameras $P_i$ and $P_j$ is defined as follows. 
We first generate a cubical voxel grid of size $32^3$ in the center of the scene with the voxel size set such that the majority of the grid is observed by all cameras in the scene.
Each point $\bx_k$, denoting the coordinates of the center of a cell in the voxel grid, is then projected to both cameras leading to a pair of projection rays $\br^i_k, \br^j_k \in \mathbb{S}^2$.
We then define the similarity 
$
s(\br^i_k, \br^j_k)
=
\delta[\pi_{P_i}(\bx_k) \in \Omega_i \land \pi_{P_j}(\bx_k) \in \Omega_j] 
(1 + \br^i_k \cdot \br^j_k)
$ 
as a dot product between the pair of rays weighted by an indicator that checks whether the projection of $\bx_k$ simultaneously lands in the rasters $\Omega_i, \Omega_j \in [0, W] \times [0, H]$ of both cameras $P_i$ and $P_j$.
The camera distance $d^\text{cam}$ is then defined as one minus the intersection-over-union of the similarities between all pairs of rays generated by each voxel grid point $\bx_k$:
 $$
 d^\text{cam}(P_i, P_j) = 1 - \frac{
 \sum_{k} s(\br^i_k, \br^j_k)
 }{
 \sum_{k} s(\br^i_k, \br^i_k) + s(\br^j_k, \br^j_k) - s(\br^i_k, \br^j_k)
 }.
 $$
Intuitively, the camera distance is proportional to the angle between the camera heading vectors adjusted by the overlap between the voxels observed by both cameras.
However, merely considering the heading vectors would not take into account the intrinsics of the cameras (focal length / principal point).
We thus devised $d^\text{cam}$ which leverages angles between projection rays, which are a function of both the intrinsics and extrinsics.

In order to understand $d^\text{cam}$, consider the following two examples: Two cameras observing the same set of voxels at a relative angle of $0.5 \pi$ would have $d^\text{cam}(P_i, P_j) \approx \frac{2}{3}$, while opposite-facing cameras would yield a maximum possible $d^\text{cam}(P_i, P_j) \approx 1$.

\paragraph{Camera difficulty bins} Each testing target camera $P^\text{tgt}$ is then assigned into one of 3 difficulty bins (\emph{easy, medium, hard}) depending on its difficulty measure $\mathcal{D}(P^\text{tgt})$.
More specifically, the easy cameras satisfy $0 \le \mathcal{D}(P^\text{tgt}) < \frac{1}{6}$,
medium $\frac{1}{6} \le \mathcal{D}(P^\text{tgt}) < \frac{1}{3}$, and hard 
$\mathcal{D}(P^\text{tgt}) \ge \frac{1}{3}$.

\section{Additional technical details}\label{s:sup_tech_details}
In this section we provide additional details of \method{\TheMethod} and of the benchmarked baseline approaches which were outlined in \cref{s:exp}.

\subsection{\textbf{\method{\TheMethod}}}

\paragraph{Source image features $\Psi$}%
The dense pixel-wise descriptor $\Psi_\text{CNN}(I^\text{src})$ (\cref{s:method_boosting_wce}) of a source image $I^\text{src}$ is a stacking of 3 types of feature tensors along the channel dimension after differentiably upsampling to a common spatial resolution $H \times W$.
The feature types are: 1) Intermediate activations extracted from the source image $I^\text{src}$ after ``layer1", ``layer2", and ``layer3" layers of the ResNet34 \cite{he15deep} network, 2) the source segmentation mask $M^\text{src}$, 3) the raw source image $I^\text{src}$.
Note that we separately map the output of each of the ResNet34 layers to a 32-dimensional feature with a 1x1 convolution followed by $\ell_2$ normalization of the feature column at every spatial location.

\paragraph{\method{\TheMethod} architecture}%
In \cref{f:sup_nerformer} we provide a more detailed visualisation of the \TheMethod architecture.

\paragraph{Rendering details}%
Similar to \method{NeRF} \cite{mildenhall2020nerf}, \method{\TheMethod} optimizes loss functions for 800 randomly sampled image rays $\br_\bu$ in each training target image.
Following \cite{mildenhall2020nerf}, \TheMethod implements a coarse and fine rendering network $f_\text{TR}$.
The former, given a ray $\br_\bu$, samples 32 points $\bx_i \in \br_\bu$ at uniform depth intervals between predefined lower and upper depth bounds.
The fine rendering network then samples 16 points on $\br_\bu$ with importance sampling from the distribution proportional to the coarse rendering weights $w_i$.

\paragraph{Training details}%
For a randomly sampled pixel $\bu$ we thus render the color $\hat I^\text{tgt}_\bu$ and an alpha value 
$\hat M^\text{tgt}_\bu = 1 - \prod_i (1 - \exp(-\Delta f_o(\bx, \bz))) \in [0, 1]$,
where the latter denotes the total amount of light absorbed by the implicit surface
($\hat M^\text{tgt}_\bu = 1$ for complete absorption).

As noted in \cref{s:method_nerformer}, the optimized loss is a sum of the RGB squared error 
$\sum_\bu \|\hat I^\text{tgt}_\bu - I^\text{tgt}_\bu\|^2$ and the segmentation binary cross entropy (BCE)
$\sum_\bu 
M^\text{tgt}_\bu \log \hat M^\text{tgt}_\bu
+ (1 - M^\text{tgt}_\bu) \log(1 - \hat M^\text{tgt}_\bu)
$.
The latter ensures that rays that do not intersect the object of interest do not terminate in the scene and vice versa.
Following  \cite{mildenhall2020nerf}, we evaluate the losses for the fine and coarse renders and optimize their sum.

\subsection{\textbf{\method{NeRF}}}
We use the implementation of \method{NeRF} \cite{mildenhall2020nerf} from PyTorch3D \cite{ravi2020pytorch3d} which closely follows the original paper.
Similar to \method{\TheMethod}, we also add to the original losses of \method{NeRF} the BCE loss between the rendered alpha mask and the ground truth target mask.
The coloring function $c_\text{MLP}$ and the opacity function $f_\text{MLP}$ have their architecture identical to the original implementation.

\subsection{SDF methods - \textbf{\method{DVR}, \method{IDR}}}

Here we detail the two baseline methods that represent shapes with signed distance fields (SDF).
We start with introducing the SDF and a method for their rendering.

\paragraph{Signed distance fields}%
While opacity functions $f_o$ represent shapes with a measure of opaqueness of 3D spatial elements,
\emph{signed distance fields}~$f_d(\bx, \bz) \in \mathbb{R}$, express the signed euclidean distance to the nearest point $\bx' \in \mathcal{S}_f$ on the implicit surface $\mathcal{S}_f$.

\paragraph{Sphere tracing (ST)} 
While EA is the most popular method for rendering opacity fields $f_o$, ST is its analogue for signed distance fields $f_d$.
Specifically, ST renders a pixel $\bu$ by seeking the minimum of the signed distance function $f_d$ on the domain of 3D points 
belonging to the ray $\br_\bu$.
ST, during its $t$-th iteration, refines the current estimate $\bx_t^{\br_\bu}$ of the ray-surface intersection by moving $\Delta_t = f_d(\bx_t^{\br_\bu}, \bz)$ units in the direction of the projection ray: $\bx_{t+1}^{\br_\bu} = \bx_{t} + \Delta_t \br_\bu$.
Upon convergence at time $T$, the rendered color 
$\hat I^{\text{tgt}}_u = c(\bx_T^{\br_\bu}, \br_\bu, \bz)$ comprises the response of the coloring function at the estimated ray-surface intersection.

\paragraph{\method{DVR}}%
natively supports our supervisory scenario and hence no alternations were required for the training protocol of both \method{DVR+AD} and \method{DVR}.
In order to implement \method{DVR+$\gamma$}, we simply convert the input coordinates to positional embeddings and adjust the number of input channels of the first layer of DVR's implicit function accordingly.
The released code \cite{niemeyer2020differentiable} supports \method{DVR+AD} so no changes were required here.
As mentioned in the paper, unfortunately, all our attempts to merge DVR with WCE lead to a non-converging model.

\paragraph{\method{IDR}}%
Similar to DVR, IDR \cite{yariv2020multiview} supports our supervisory setup by default.
In order to implement \method{IDR+AD}, we append the latent code $\bz$ to the positional embeddings that are input to the implicit function, and we adjust the number of input channels of the first layer of the implicit function accordingly.
\method{IDR} already takes as input the positional embeddings $\gamma$, so the extension \method{IDR+$\gamma$} does not apply here.
As mentioned in the main paper, we could not obtain a converging version of \method{IDR+WCE}.

\subsection{\textbf{\method{SRN}}}%
Here, we first give a brief overview of the learned SRN raymarcher, followed by describing the WCE extension of SRN.

\paragraph{Neural raymarching} Contrasted to the explicit formulations of sphere-tracing or EA, recently, SRN \cite{sitzmann19scene} proposed to learn to march along the projection rays with a recurrent deep network.
Similar to sphere-tracing, SRN decides at iteration $t$ on the length of the raymarching step $\Delta_t$ by evaluating a function at the current intersection estimate $\bx_t^{\br_u}$.
However, instead of querying the SDF, SRN utilizes an LSTM \cite{hochreiter97long} cell 
$f_\text{LSTM}(\bx_t^{\br_\bu}, \br_\bu, \bz, \bh_t) = (\Delta_t, \bh_{t+1})$ which is additionally conditioned on the ray direction $\br_\bu$ and a temporal hidden state $\bh_t$. 
In this manner, the raymarcher adapts the step-size prediction based on the past marching observations.

\paragraph{\method{SRN+WCE}} The WCE exension of SRN is straightforwardly implemented by replacing the iterative invocation of the global-encoding-conditioned implicit 
$f_\text{LSTM}(
\bx_t^{\br_\bu}, \br_\bu, \bz_\text{global}, \bh_t
)
$
of the SRN's raymarcher with the WCE-conditioned implicit 
$f_\text{LSTM}(
\bx_t^{\br_\bu}, 
\br_\bu, 
\bz_\text{WCE}^\star(\bx_t^{\br_\bu}, \{I_i^\text{src}\}, \{P_i^\text{src}\}),
\bh_t
)
$
This way, the learned raymarcher can ``tap"" into the source views during every iteration to receive a more direct triangulation signal. 
As apparent from \cref{t:results_mscene,t:results_mscene_full}, our WCE extension of SRN provides a very strong baseline that in fact achieves the best depth prediction performance.

\paragraph{Mask prediction} The learned raymarcher of the original version of SRN does not render an alpha mask of the foreground object.
In order to enable the latter, we extend the last layer of the SRN's coloring function $c$ with an additional channel that is terminated with a sigmoid activation and represents the alpha value of the corresponding pixel $\bu$.
This channel is then supervised by minimizing the DICE coefficient between its output and the ground truth segmentation masks $M^\text{tgt}$.

\subsection{\textbf{\method{P3DMesh}}}
As mentioned in the paper, \method{P3DMesh} \cite{ravi2020pytorch3d} deforms an initial spherical mesh template with a fixed topology with a series of convolutions on the mesh graph.
As in \cite{wang2018pixel2mesh}, the graph convolutions accept features sampled from the source images at the 2D projections of the mesh vertices.
Since P3DMesh supports conditioning only on a single-source-view, we extend to the multi-view setting by averaging over the per-vertex features sampled from each of the source views.
Furthermore, note that the implementation in \cite{ravi2020pytorch3d} differentiably renders the mesh with a memory-efficient version of the Soft Rasterizer \cite{liu19soft}.
The training protocol, including the employed losses and their weighting, closely follows \cite{ravi2020pytorch3d}.

\subsection{Neural Volumes (\method{NV})} %
Neural Volumes \cite{lombardi2019neural} is a method that represents implicit surfaces as voxel grids.
In what follows, we first briefly describe voxel grids, their specific implementation in \method{NV}, and its extension with warp-conditioned embedding (\method{NV+WCE}).

\paragraph{Voxel grids}%
While MLPs can label an arbitrary element of the 3D domain, 
a voxel grid can be seen as an implicit surface restricted to a subset of $\mathbb{R}^3$ which is uniformly subdivided to a lattice $V(\bz) \in \mathbb{R}^{R^3}$ of $R^3; R \in \mathbb{N}_+$ cuboid elements of the same size.
Note that the lattice $V(\bz)$ is a function (typically a 3D deconvnet) of $\bz$ which allows for representing different 3D shapes.
The implicit function $f_\text{voxel}(\bx, \bz) = \zeta(V(\bz), \bx)$ is then evaluated by sampling $V(\bz)$ at the corresponding world coordinate $\bx$, with a grid-sampling function $\zeta: \mathbb{R}^{R^3} \times \mathbb{R}^3 \mapsto \mathbb{R}^{\text{dim}(f)}$, such as trilinear interpolation.
Voxel grids also admit coloring via a volume $C(\bz) \in \mathbb{R}^{3 \times R^3}$ which can be sampled in an analogous manner.

\paragraph{Neural Volumes}%
A notable voxel-grid-based method is Neural Volumes \cite{lombardi2019neural}, which proposed an improved sampling function 
$\zeta_\text{warp}(\zeta(W(\bz), \bx) + \bx, V(\bz))$ which refines the sampling location $\bx$ with an offset vector
$\zeta(W(\bz), \bx) \in \mathbb{R}^3$ sampled from a warping lattice $W(\bz) \in \mathbb{R}^{R^3}$.
Here both $W$ and $V$ are implemented as a 3D deconvolutional network.

\paragraph{\method{NV} and \method{NV+AD}}%
\method{NV+AD} is in fact the vanilla version of \cite{lombardi2019neural} whose 3D deconvnets $V$ and $W$ accept the scene-specific latent code $\bz_\text{scene}(\text{sequence}_\text{ID})$ (described in \cref{s:exp_multi_scene}).
The "overfitting" version of \method{NV} from \cref{t:results_sscene} is a special case of \method{NV+AD} with a single latent code.

\paragraph{\method{NV+WCE}}%
The WCE extension of Neural Volumes (\method{NV+WCE}) appends the WCE to the feature of each voxel after the second 3D deconvolution layer of the 3D convnets $V$ and $W$.
Here, the WCE of a voxel is generated by expressing the world coordinate $\bx_{V_i}$ of the center of the correspoding voxel $V_i$ and calculating the aggregate WCE $\bz_\text{WCE}^\star(\bx_{V_i}, \{I_i^\text{src}\}, \{P_i^\text{src}\})$ for a set of source views $\{I_i^\text{src}\}$ and their cameras $\{P_i^\text{src}\}$.
Note that a similar approach has been proposed in \cite{kar2017learning}.

\paragraph{Training}%
All versions of \method{NV} optimize the losses from \cite{lombardi2019neural} with the original weights.
Furthermore, we exploit the known ground truth segmentation masks and minimize the binary cross entropy between the alpha mask returned by the raymarcher of \method{NV} and the ground truth mask $M^\text{tgt}$.

\subsection{Implicit Point Cloud (\method{IPC})}%
As mentioned in \cref{s:exp}, \method{IPC} represents shapes by converting a point cloud to an implicit function which is later rendered with EA raymarching.

Formally, let a point cloud $\mathcal{P}(\bz) = \{ \bx_i \}_{i=1}^{N_\text{pts}}$ be an $N_\text{pts}$-sized unordered set of points, 
where $\mathcal{P}$ is a point cloud predictor (detailed later in this section) which accepts the latent code $\bz$.
$\mathcal{P}(\bz)$ then admits an occupancy function $f_{\mathcal{P} \epsilon}$ defined as follows:
$$
f_{\mathcal{P} \epsilon}(\bx', \bz) = 
\delta[\| \text{NN}_{\mathcal{P}(\bz)}(\bx') - \bx'\| < \epsilon],
$$ 
where 
$\text{NN}_{\mathcal{P}(\bz)}(\bx') = \argmin_{\bx \in \mathcal{P}(\bz)} \| \bx - \bx' \|$
returns the nearest point from the point cloud $\mathcal{P}(\bz)$ to the query point $\bx'$.
Intuitively, $f_{\mathcal{P} \epsilon}$  yields zero everywhere except within an $\epsilon$ neighborhood of each point cloud point $\bx_i \in \mathcal{P}(\bz)$, where $f_{\mathcal{P}}$ yields 1.
As we describe later, anchoring the implicit function on the set of cloud points allows for faster and more memory-efficient EA raymarching than in the case of the neural implicit occupancy $f_\text{MLP}$ (described in \cref{s:method_shape_representations}).

In order to color the implicit point cloud, we define its coloring function $c_\text{IPC}$:
$$
c_\text{IPC}(\bx', \br, \bz) 
=
c_\text{MLP}(\text{NN}_{\mathcal{P}(\bz)}(\bx'), \br, \bz).
$$
Here $c_\text{IPC}$ attaches to an arbitrary point $\bx'$ the response of the coloring MLP $c_\text{MLP}$ at $\bx'$'s nearest point cloud neighbor $\text{NN}_{\mathcal{P}(\bz)}(\bx')$.

\paragraph{Rendering \method{IPC}}%
\method{IPC} is rendered efficiently with the PyTorch3D point cloud renderer \cite{ravi2020pytorch3d,wiles2020synsin}.
More specifically, given a target camera $P^\text{tgt}$, each point from the predicted point cloud $\mathcal{P}(\bz)$ is projected to the camera plane to form a set of 2D projections
$\{\pi_{P^\text{tgt}}(\bx_i) | \bx_i \in \mathcal{P}(\bz)\}$.
For each pixel coordinate $\bu \in \{1, ..., W\} \times \{1, ..., H\}$ in the rendering lattice of the target render $\hat I^\text{tgt} \in \mathbb{R}^{3 \times H \times W}$, the renderer records the ordered set
\begin{align*}
\mathsf{\Pi}_\epsilon^\bu(\mathcal{P}(\bz))
= \big(
\bx_i |&
\bx_i \in \mathcal{P}(\bz);
\|\pi_{P^\text{tgt}}(\bx_i) - \bu\|
\le \epsilon \mathsf{f}_{P^\text{tgt}}; \\
& d_{P^\text{tgt}}(\bx_i) \le d_{P^\text{tgt}}(\bx_{i + 1})
\big),
\end{align*}
of point cloud points $\bx_i \in \mathcal{P}(\bz)$ whose 2D projections $\pi_{P^\text{tgt}}(\bx_i)$ land within the $\epsilon \mathsf{f}_{P^\text{tgt}}$ distance from the pixel $\bu$, and which is ordered by the depth $d_{P^\text{tgt}}(\bx_i)$ of each point in the target camera $P^\text{tgt}$.
$\mathsf{f}_{P^\text{tgt}} \in \mathbb{R}$ is the focal length of the target camera $P^\text{tgt}$.

Intuitively, $\mathsf{\Pi}_\epsilon^\bu(\mathcal{P}(\bz))$ denotes the set of point cloud points whose $\epsilon$ neighborhoods are intersected by the rendering ray $\br_\bu$ emitted from pixel $\bu$.
Note that this is an approximation: comparing the 2D camera-plane distance $\|\pi_{P^\text{tgt}}(\bx_i) - \bu\|$ 
to the constant $\epsilon \mathsf{f}_{P^\text{tgt}}$ corresponds to orthographic projections of the point neighborhoods, whereas our cameras are perspective.
However, the orthographic approximation is mild in our case, since the distance of the point cloud points from the camera is relatively large compared to its focal length.

The EA raymarching then takes the set of $\bu$'s 3D points $\mathsf{\Pi}_\epsilon^\bu(\mathcal{P}(\bz))$ in order to render the color $\hat I^\text{tgt}_\bu \in \mathbb{R}^3$:
$$
\hat I^\text{tgt}_\bu(\br_\bu, \bz) = \sum_{\bx_i \in \mathsf{\Pi}_\epsilon^\bu(\mathcal{P}(\bz))}
w_i(\bx_i, \bz, \bu) c_\text{IPC}(\bx_i, \br_\bu, \bz).
$$
For \method{IPC}, the weight $w_i(\bx_i, \bz, \bu) = \left(\prod_{j=0}^{i-1} T_j^\text{IPC}(\bx_i, \bz, \bu)\right) \left(1 - T_i^\text{IPC}(\bx_i, \bz, \bu)\right)$ is the product of emission and absorption functions with the
transmission term $T_i^\text{IPC}$ defined as
$$
T_i^\text{IPC}(\bx_i, \bz, \bu) = 
\underbrace{f_{\mathcal{P} \epsilon}(\bx_i, \bz)}_{=1}
\frac{\|\bu - \pi_{P^\text{tgt}}(\bx_i)\|}
{\epsilon \mathsf{f}_{P^\text{tgt}}}
,
$$
which approximately measures the amount of light transmitted through the spherical $\epsilon$ neigborhood of a point $\bx_i$ which intersects the projection ray $\br_\bu$.
To demonstrate this, observe that for a pixel $\bu^\text{intersect}=\pi_{P^\text{tgt}}(\bx_i)$ which coincides with the projection of the 3D point $\bx_i$, the transmission $T_i^\text{IPC}(\bx_i, \bz, \bu^\text{intersect})=0$, \ie no light is transmitted through $\bx_i$ and the corresponding color 
$c_\text{IPC}(\bx_i, \br_{\bu^\text{intersect}}, \bz)$ is fully rendered.
On the contrary, for a pixel $\bu^\text{outside} = \pi_{P^\text{tgt}}(\bx_i + \epsilon)$ outside the epsilon neighborhood, the unit transmission $T_i^\text{IPC}(\bx_i, \bz, \bu^\text{outside})=1$ signifies that all light passes through and the point's color is ignored during rendering.
Note that the above equation is very similar to the top-k point cloud rasterizer of SinSyn \cite{wiles2020synsin}.

\paragraph{Point cloud predictor $\mathcal{P}(\bz)$} The point cloud predictor $\mathcal{P}(\bz)$ is the same for both \method{IPC+AD} and \method{IPC+WCE}.
More specifically, $\mathcal{P}(\bz) = \{ \bar \bx_i + o_\text{MLP}(\bar \bx_i, \bz) \}_{i=1}^{N_\text{pts}}$ offsets a fixed set of template points 
$\mathcal{\bar P} = \{ \bar \bx_i \}_{i=1}^{N_\text{pts}}$ with an offset function 
$o: \mathbb{R}^3 \times \mathbb{R}^{D_\bz} \mapsto \mathbb{R}^3$ implemented as an MLP with the same architecture as $f_\text{MLP}$.
Therefore, $o$ alters the template point cloud to match a specific shape given its latent shape code $\bz$.

\paragraph{\method{IPC+AD} and \method{IPC+WCE}}
For \method{IPC+AD}, the offset function $o$ accepts the video-specific latent code $\bz_\text{scene}(\text{sequence}_\text{ID})$ described in \cref{s:exp_multi_scene}, 
while for \method{IPC+WCE}, $o$ takes as input the aggregate warp-conditioned embedding 
$\bz_\text{WCE}^\star(\bar \bx_i, \{I_i^\text{src}\}, \{P_i^\text{src}\})$ 
evaluated at each template point $\bar \bx_i \in \mathcal{\bar P}$.
Finally, the single-scene version, abbreviated simply as \method{IPC} in \cref{t:results_sscene}, is a special case of \method{IPC+AD} with $\bz_\text{scene}(\text{sequence}_\text{ID}) := \mathbf{0}$ set to a constant zero vector.

\begin{table*}[th]
\small\centering%
\begin{tabular}{l|c|p{0.75\linewidth}}\toprule
Metric & Domain & Description \\ \midrule
$BA_\text{final cost}$ & $\mathbb{R}$ & The final value of the Bundle Adjustment (BA) cost function. \\
$BA_\text{termination}$ & $\{0, 1\}$ & The termination state of BA (converged/not converged). \\
$\mu_\text{det score}$ & $[0, 1]$ & An average over per-frame detection scores of the PointRend object detector. \\
$\mu_\text{perc detected}$ & $[0, 100]$ & Percentage of frames in which the category of interest is detected with PointRend. \\
$N_\text{cameras}$ & $\mathbb{N}$ & The number of cameras registered during BA. \\
$N_\text{sparse pts}$ & $\mathbb{N}$ & Number of points in the sparse point cloud. \\
$PCL^\text{render}_{\ell^\text{depth}}$ & $\mathbb{R}$ &
The average $\ell_1$ depth error between the renders of the fused pointcloud $\mathcal{P}(\mathcal{V})$ into each camera $P_i$ of a video $\mathcal{V}$ and the corresponding dense depth map $P_i$. \\
$PCL^\text{render}_{\ell^\text{rgb}}$ & $\mathbb{R}$ &
The average $\ell_1$ RGB error between the renders of the fused pointcloud $\mathcal{P}(\mathcal{V})$ into each camera $P_i$ of a video $\mathcal{V}$ and the corresponding frame $I_i$. \\
$PCL^\text{render}_{IoU}$ & [0, 1] &
The average Jaccard Index between the renders of the fused point cloud $\mathcal{P}(\mathcal{V})$ into each camera $P_i$ of a video $\mathcal{V}$ and the corresponding PointRend segmentation $M_i$. \\
$PCL^\text{direction cover}$ & $\mathbb{N}$ &
Measures the coverage of the views of the point cloud $\mathcal{P}(\mathcal{V})$ with the number of occupied bins in the azimuth/elevation map of projection rays corresponding to each dense point cloud point $\bx_j$ and a camera $P_\text{i}$.\\\bottomrule
\end{tabular}
\caption{\textbf{The list of SfM and point-cloud reconstruction metrics} that serve as a set of features for training the active-SVM that labels camera and reconstruction quality.\label{t:sup_reconmetrics}}
\end{table*}

\paragraph{Training}%
All versions of \method{IPC} optimize the MSE between the rendered image $\hat I^\text{tgt}$ and the ground truth colors $I^\text{tgt}$.
Furthermore, we make use of the ground truth segmentation masks and minimize the Chamfer distance between the set of 2D projections of the predicted point cloud points $\mathcal{P}(\bz)$, and the 2D points of the ground truth segmentation mask \cite{li2020self}.
Note that a standard segmentation loss, such as DICE \cite{sudre2017generalised} or Binary Cross Entropy between the rendered alpha mask and the ground truth segmentation mask, do not apply here.
This is because the gradients generated by the alpha mask renders of \method{IPC} are not well-defined and do not lead to convergence.

\section{3D annotations with Human-in-the-loop}\label{s:sup_human_il}

In \cref{s:data_3dgt}, we outlined the process of annotating the AMT-collected videos with 3D ground truth.
Here, we further detail the semi-automated process of labelling the quality of camera tracking and the 3D dense point cloud of the captured videos (Paragraph 4 in \cref{s:data_3dgt}).

We initialize the process by annotating an initial set of several hundreds of reconstructions with a binary label "accurate / inaccurate" by visually inspecting both the camera tracks 
$(P_i | P_i \in R^{4 \times 4})_{i=1}^{N_I}$
and the scene point cloud $\mathcal{P}(\mathcal{V})$.
From each video, we then extract various metrics that are indicative of the reconstruction quality such as a per-pixel RGB and depth error of the rendered point cloud, the number of registered cameras, final bundle adjustment energy etc.
The full set of metrics is outlined in \cref{t:sup_reconmetrics}.
We then train a binary Support Vector Machine (SVM \cite{cortes95support-vector}) with an RBF kernel that regresses the binary label given the reconstruction metrics as input.

Afterwards, the trained SVM classifies all previously unlabelled videos.
In line with the uncertainty principle \cite{tong01support}, we manually annotate a subset of previously unlabelled samples that are the closest to the SVM decision boundary.
We further correct significant classification errors by inspecting the highest/lowest scoring samples.
In this manner, we alternate between SVM training and manual annotation until 1.5k labels are collected (8 \% of the whole dataset). 

In order to validate the SVM's performance, we conduct a 5-fold cross-validation on the set of annotated videos.
The cross-validation indicates that the SVM has 90\% and 78\% accuracy for classifying the camera tracking and point cloud quality respectively.


\else

\clearpage
\appendix

{\small%
\bibliography{refs,vedaldi_general,vedaldi_specific}
}

\fi

\end{document}